\documentclass[10pt,twocolumn,letterpaper]{article}

\usepackage[pagenumbers]{cvpr} 

\usepackage{times}
\usepackage{epsfig}
\usepackage{graphicx}
\usepackage{amsmath}
\usepackage{amssymb}

\usepackage{bbm}
\usepackage{bm}
\usepackage{xspace}
\usepackage[utf8]{inputenc} 
\usepackage{url}            
\usepackage{booktabs}       
\usepackage{amsfonts}       
\usepackage{nicefrac}       
\usepackage{microtype}      
\usepackage{color}
\usepackage{algorithm}
\usepackage{algorithmic}
\usepackage{array}
\usepackage{multirow}
\usepackage{enumitem}
\usepackage{makecell}
\usepackage{gensymb}         
\usepackage{mathtools}
\usepackage{dblfloatfix}
\usepackage{pifont}
\usepackage[permil]{overpic}

\usepackage[accsupp]{axessibility}

\usepackage[pagebackref,breaklinks,colorlinks]{hyperref}

\definecolor{red}{rgb}{1,0,0}
\definecolor{slateblue}{rgb}{0.7,0.35,0.9}
\definecolor{green}{rgb}{0,1,0}
\definecolor{mahogany}{rgb}{0.75, 0.25, 0.0}
\definecolor{purple}{rgb}{0.6, 0, 0.6}
\definecolor{darkpurple}{rgb}{0.3, 0, 0.3}
\definecolor{darkgreen}{rgb}{0, 0.4, 0}
\definecolor{frenchblue}{rgb}{0.0, 0.45, 0.73}
\definecolor{blue}{rgb}{0,0,1}
\definecolor{goldenrod}{rgb}{0.65, 0.45, 0.03}
\definecolor{gray}{rgb}{0.6,0.6,0.6}
\definecolor{gold}{rgb}{1.0, 0.874, 0}
\definecolor{silver}{rgb}{0.67,0.67,0.67}
\definecolor{brown}{rgb}{0.8, 0.678, 0.4}

\usepackage[capitalize]{cleveref}
\crefname{section}{Sec.}{Secs.}
\Crefname{section}{Section}{Sections}
\Crefname{table}{Table}{Tables}
\crefname{table}{Tab.}{Tabs.}

\newtheorem{definition}{Definition}
\newtheorem{problem}{Problem}

\def\cvprPaperID{1974} 
\def\confName{CVPR}
\def\confYear{2023}

\newcommand{\ourmethod}{F$^2$-NeRF }

\begin{document}

\title{F$^{2}$-NeRF: Fast Neural Radiance Field Training with Free Camera Trajectories}

\author{
Peng Wang$^{1,2 *}$
\qquad
Yuan Liu$^{1}\thanks{Equal contribution.}$
\qquad
Zhaoxi Chen$^{2}$
\qquad
Lingjie Liu$^{3}$ \\
\qquad
Ziwei Liu$^{2}$
\qquad
Taku Komura$^{1}$
\qquad
Christian Theobalt$^{3}$
\qquad
Wenping Wang$^{4}$
\vspace{0.1cm}\\
{\normalsize $^{1}$ The University of Hong Kong \quad $^{2}$ S-Lab, Nanyang Technological University}
\\
{\normalsize \quad $^{3}$ Max Planck Institute for Informatics \quad $^{4}$ Texas A\&M University}
}
\maketitle

\begin{abstract}

This paper presents a novel grid-based NeRF called \textbf{F${^2}$-NeRF (Fast-Free-NeRF)} for novel view synthesis, which enables arbitrary input camera trajectories and only costs a few minutes for training. Existing fast grid-based NeRF training frameworks, like Instant-NGP, Plenoxels, DVGO, or TensoRF, are mainly designed for bounded scenes and rely on space warping to handle unbounded scenes. Existing two widely-used space-warping methods are only designed for the forward-facing trajectory or the 360$^\circ$ object-centric trajectory but cannot process arbitrary trajectories. In this paper, we delve deep into the mechanism of space warping to handle unbounded scenes.
Based on our analysis, we further propose a novel space-warping method called perspective warping, which allows us to handle arbitrary trajectories in the grid-based NeRF framework.
Extensive experiments demonstrate that \ourmethod is able to use the same perspective warping to render high-quality images on two standard datasets and a new free trajectory dataset collected by us. Project page: \url{https://totoro97.github.io/projects/f2-nerf}.

\end{abstract}

\section{Introduction}
\label{sec:intro}

\begin{figure}[!t]
  \includegraphics[width=\linewidth]{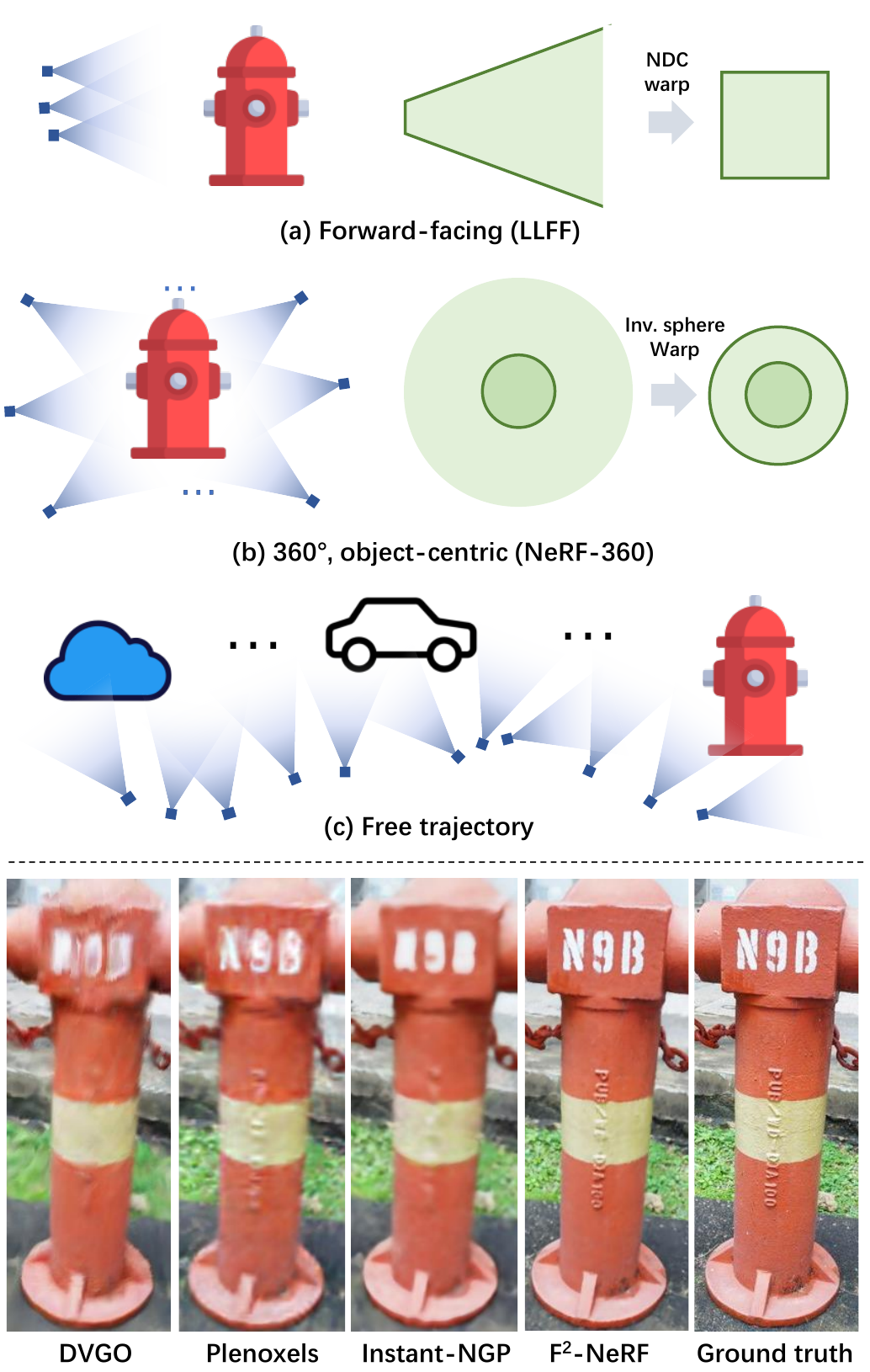} \caption{Top: (a) Forward-facing camera trajectory. (b) 360$^\circ$ object-centric camera trajectory. (c) Free camera trajectory. In (c), the camera trajectory is long and contains multiple foreground objects, which is extremely challenging. Bottom: Rendered images of the state-of-the-art fast NeRF training methods and \ourmethod on a scene with a free trajectory.}
  \label{fig:teaser}
  \vspace{-1em}
\end{figure}

The research progress of novel view synthesis has advanced drastically in recent years since the emergence of the Neural Radiance Field (NeRF)~\cite{MildenhallSTBRN20,tewari2022advances}. Once the training is done, NeRF is able to render high-quality images from novel camera poses. The key idea of NeRF is to represent the scene as a density field and a radiance field encoded by Multi-layer Perceptron (MLP) networks, and optimize the MLP networks with the differentiable volume rendering technique. Though NeRF is able to achieve photo-realistic rendering results, training a NeRF takes hours or days due to the slow optimization of deep neural networks, which limits its application scopes. 

Recent works demonstrate that grid-based methods, such as Plenoxels~\cite{YuFTCR22}, DVGO~\cite{SunSC22}, TensoRF~\cite{ChenXGYS22}, and Instant-NGP~\cite{mueller2022instant}, enable fast training a NeRF within a few minutes. However, the memory consumption of such grid-based representations grows in cubic order with the size of the scene. Though various techniques, such as voxel pruning~\cite{YuFTCR22,SunSC22}, tensor decomposition~\cite{ChenXGYS22} or hash indexing~\cite{mueller2022instant}, are proposed to reduce the memory consumption, these methods still can only process bounded scenes when grids are built in the original Euclidean space.

To represent unbounded scenes, a commonly-adopted strategy is to use a space-warping method that maps an unbounded space to a bounded space~\cite{MildenhallSTBRN20, BarronMVSH22, ZhangRSK20}.
There are typically two kinds of warping functions.
(1) For forward-facing scenes (Fig.~\ref{fig:teaser}~(a)), the Normalized Device Coordinate (NDC) warping is used to map an infinitely-far view frustum to a bounded box by squashing the space along the z-axis~\cite{MildenhallSTBRN20}; (2) For 360$^\circ$ object-centric unbounded scenes (Fig.~\ref{fig:teaser}~(b)), the inverse-sphere warping can be used to map an infinitely large space to a bounded sphere by the sphere inversion transformation~\cite{BarronMVSH22, ZhangRSK20}. Nevertheless, these two warping methods assume special camera trajectory patterns and cannot handle arbitrary ones. In particular, when a trajectory is long and contains multiple objects of interest, called \textit{free trajectories}, as shown in Fig.~\ref{fig:teaser}~(c), the quality of rendered images degrades severely.

The performance degradation on free trajectories is caused by the imbalanced allocation of spatial representation capacity. Specifically, when the trajectory is narrow and long, many regions in the scenes are empty and invisible to any input views. However, the grids of existing methods are regularly tiled in the whole scene, no matter whether the space is empty or not. Thus, much representation capacity is wasted on empty space. 
Although such wasting can be alleviated by using the progressive empty-voxel-pruning~\cite{YuFTCR22, SunSC22}, tensor decomposition~\cite{ChenXGYS22} or hash indexing~\cite{mueller2022instant}, it still causes blurred images due to limited GPU memory. Furthermore, in the visible spaces, multiple foreground objects in Fig.~\ref{fig:teaser}~(c) are observed with dense and near input views while background spaces are only covered by sparse and far input views. In this case, for the optimal use of the spatial representation of the grid, dense grids should be allocated for the foreground objects to preserve shape details and coarse grids should be put in background space.
However, current grid-based methods allocate grids evenly in the space, causing the inefficient use of the representation capacity.

To address the above problems, we propose \ourmethod (Fast-Free-NeRF), the first fast NeRF training method that accommodates free camera trajectories for large, unbounded scenes. Built upon the framework of Instant-NGP~\cite{mueller2022instant}, \ourmethod can efficiently be trained on unbounded scenes with diverse camera trajectories
and maintains the fast convergence speed of the hash-grid representation.

In \ourmethod, we give the criterion on a proper warping function under an arbitrary camera configuration. Based on this criterion, we develop a general space-warping scheme called the \textit{perspective warping} that is applicable to arbitrary camera trajectories. The key idea of perspective warping is to first represent the location of a 3D point ${\bf p}$ by the concatenation of the 2D coordinates of the projections of ${\bf p}$ in the input images and then map these 2D coordinates into a compact 3D subspace space using Principle Component Analysis (PCA)~\cite{wold1987principal}. We empirically show that the proposed perspective warping is a generalization of the existing NDC warping~\cite{MildenhallSTBRN20} and the inverse sphere warping~\cite{ZhangRSK20,BarronMVSH22} to arbitrary trajectories in a sense that the perspective warping is able to handle arbitrary trajectories while could automatically degenerate to these two warping functions in forward-facing scenes or 360$^\circ$ object-centric scenes. In order to implement the perspective warping in a grid-based NeRF framework, we further propose a space subdivision algorithm to adaptively use coarse grids for background regions and fine grids for foreground regions.

We conduct extensive experiments on the unbounded forward-facing dataset, the unbounded 360$^\circ$ object-centric dataset, and a new unbounded free trajectory dataset. The experiments show that \ourmethod uses the same perspective warping to render high-quality images on the three datasets with different trajectory patterns. On the new Free dataset with free camera trajectories, our method outperforms baseline grid-based NeRF methods,while only using $\sim$12 minutes on training on a 2080Ti GPU.

\section{Related Works}
\textbf{Novel view synthesis.}
Novel view synthesis (NVS) aims to synthesize novel view images from input posed images. The NVS problem has been extensively studied with lumigraph~\cite{BuehlerBMGC01,GortlerGSC96} and light field functions~\cite{DavisLD12,LevinD10} to directly interpolate input images. To improve the quality of the synthesized image, many methods resort to an explicit 3D reconstruction of the scene via meshes~\cite{DebevecTM96,ThiesZN19,WaechterMG14,WoodAACDSS00}, voxels~\cite{HeCJS20,LombardiSSSLS19,LombardiSSZSS21,SitzmannTHNWZ19}, point clouds~\cite{aliev2020neural,xu2022point}, depth maps~\cite{DhamoTLNT19,ShadeGHS98,ShihSKH20,TulsianiTS18}, and multi-plane images (MPI)~\cite{FlynnBDDFOST19,LiXDS20,MildenhallSCKRN19,SrinivasanTBRNS19,TuckerS20,ZhouTFFS18}, and then synthesize novel images with the help of these 3D reconstructions. \ourmethod also aims to solve the NVS task but with a neural representation.

\textbf{Neural scene representations.}
Since the emergence of NeRF~\cite{MildenhallSTBRN20, BarronMTHMS21, tancik2023nerfstudio}, there have been intensive studies on neural representations for the tasks of novel view synthesis~\cite{MildenhallSTBRN20,SitzmannZW19,tewari2022advances}, relighting~\cite{BossBJBLL21,munkberg2022extracting,zhang2021physg,ZhangSDDFB21}, generalization to new scenes~\cite{YuYTK21,WangWGSZBMSF21,LiuPLWWTZW21,ChenXZZXYS21,wang2022generalizable,suhail2022generalizable}, shape representation~\cite{martel2021acorn, takikawa2021neural, mueller2022instant}, and multi-view reconstruction~\cite{yariv2020multiview, wang2021neus, OechslePG21, yariv2021volume}. The representation can be either totally neural networks~\cite{MildenhallSTBRN20, sitzmann2020implicit, fathony2020multiplicative, lindell2022bacon, ramasinghe2022beyond}, or hybrid parametric encodings with space subdivisions~\cite{takikawa2021neural,martel2021acorn,mueller2022instant,LiuGLCT20} for efficient training and inference. \ourmethod also subdivides the scene for flexible space warping and uses the hybrid neural scene representation~\cite{mueller2022instant} for fast training and high-quality rendering.

\textbf{Fast NeRF training with space warping.}
Recent works show that the training of NeRF can be accelerated significantly with grid-based representations~\cite{YuFTCR22,ChenXGYS22,SunSC22,mueller2022instant}. Instead of using a huge MLP network to predict the density and color, Plenoxels~\cite{YuFTCR22} directly store the density values and colors on a voxel grid. Instant-NGP~\cite{mueller2022instant}, TensoRF~\cite{ChenXGYS22} and DVGO~\cite{SunSC22} construct a feature grid and the density and compute the density and the color for a specific point from an interpolated feature vector using a tiny MLP network. However, these grids are regularly constructed in an axis-aligned manner and require additional space warping to represent unbounded scenes. There are two kinds of existing space warping functions, the NDC warping~\cite{MildenhallSTBRN20} for the forward-facing scenes and the inverse sphere warping~\cite{BarronMVSH22,BarronMTHMS21,ZhangRSK20} for the 360$^\circ$ object-centric scenes. Both of these warping functions cannot handle long and narrow trajectories. In \ourmethod, we propose a novel perspective warping to enable these fast grid-based methods to process arbitrary camera trajectories.

\textbf{Large-scale Neural Radiance Fields.}
Recent works~\cite{turki2022mega,xiangli2022bungeenerf,tancik2022block} managed to reconstruct the radiance field on a large-scale scene by decomposing the scene into blocks and separately training different NeRFs for different blocks. \ourmethod aims at small-scale scenes with arbitrary camera trajectories, which has the potential to serve as the backbone NeRF of a single block in large-scale NeRFs.

\begin{figure}[!b]
  \includegraphics[width=\linewidth]{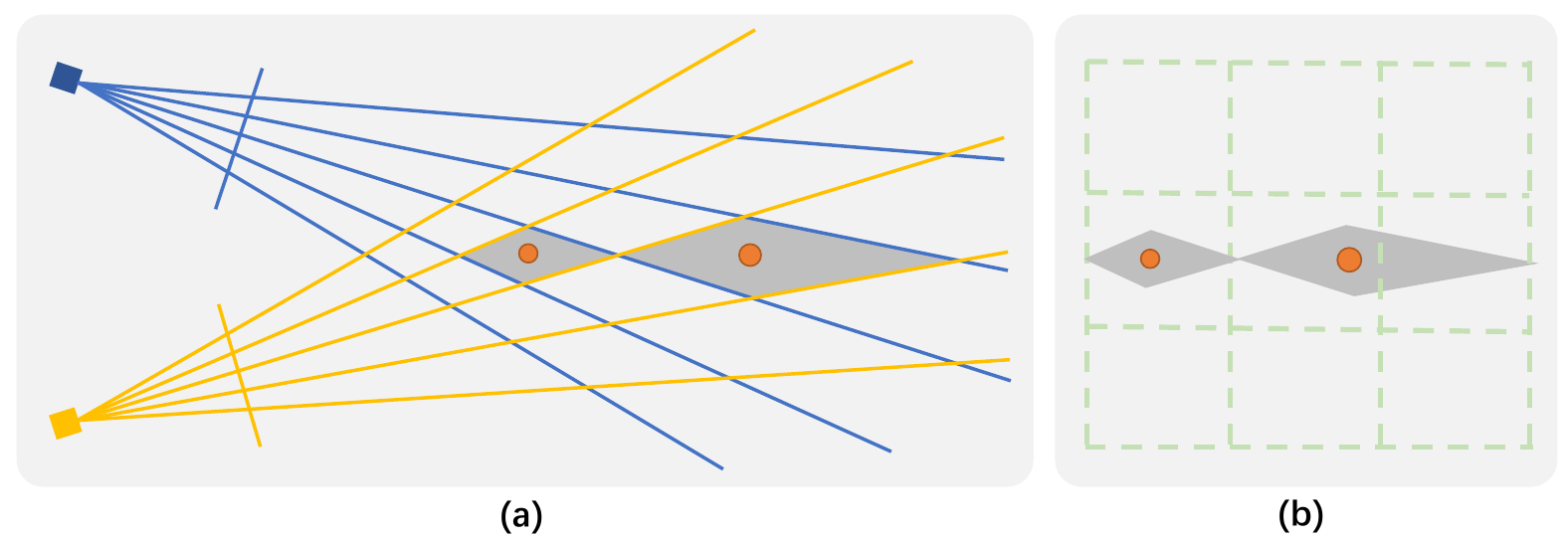} \caption{{\bf A 2D example.} (a) The gray regions at the orange points align with the image resolutions. (b) Axis-aligned grids in the original Euclidean space are not aligned with the camera rays.}
  \label{fig:method_2d_warp}
\end{figure}

\section{Our Approach}
Given a set of images $\{\mathcal{I}_i \}$ with arbitrary but known poses in an unbounded scene, the goal of \ourmethod is to reconstruct a radiance field of the scene for the novel view synthesis task. In the following, we first give an overview of \ourmethod.

\subsection{Overview}
\label{sec:pre}
In order to build a grid-based neural representation in an unbounded scene, a space warping function $F(\mathbf{x})$ is introduced to warp the unbounded space to a bounded region. In Sec.~\ref{sec:per}, we first analyze the mechanism of space warping and propose a novel perspective warping method. The proposed perspective warping subdivides whole the space under consideration into small regions, as shown in Sec.~\ref{sec:sub}. Then, on the warp space, we build a grid-based neural representation in Sec.~\ref{sec:rep}. 
Based on the built representation, we adopt the volume rendering~\cite{MildenhallSTBRN20} to render novel-view images. 

\textbf{Volume rendering with perspective warping}.
In order to render the color $\hat c$ for a pixel, we first apply a novel point sampling strategy, called the \textit{perspective sampling} in Sec.~\ref{sec:method_sampling}, to sample points $\mathbf{x}_i$ on the camera ray emitting from the pixel.
Then, these sampled points are warped by the perspective warping 
to the warp space, and the density $\sigma_i$ and the color $c_i$ on sampled points are computed from the neural representation built on the warp space.
Finally, we composite the colors to compute the pixel color $\hat c$ by
\begin{equation}
\hat c = \sum_i^{}{T_i\alpha_i c_i},
\end{equation}
where $T_i = \prod_{j=0}^{i-1} (1 - \alpha_j)$ is the accumulated transmittance and $\alpha_i = 1 - \exp(-\delta_i\sigma_i)$ is the opacity of the point.

\subsection{Perspective warping}
\label{sec:per}
It has been demonstrated that space warping functions, such as the NDC warping~\cite{MildenhallSTBRN20} and the inverse sphere warping~\cite{ZhangRSK20,BarronMVSH22}, are effective for rendering unbounded scenes.
In this section, we start with a 2D intuitive analysis of why a space warping method is effective.

{\bf 2D analysis.} First, let us consider a simple case in the 2D space as shown in Fig.~\ref{fig:method_2d_warp}~(a). In this setting, two 2D cameras project the points from 2D space onto their 1D image planes. Consider the two orange points in the figure. The gray rhombuses are the irregular grids formed by camera rays and the gray regions are the smallest distinguishable region due to the limited resolution of the two cameras. 
However, a vanilla grid-based representation consists of axis-aligned regular grids as shown in Fig.~\ref{fig:method_2d_warp}~(b), which is not aligned with the gray rhombuses.
Moreover, such misalignment becomes more severe as the distance from the cameras increases.
The key requirement for space warping is that we need to warp the original Euclidean space and build axis-aligned grids in the warp space so that these grids are aligned with the camera rays. 
Clearly, in this 2D case, a proper warping function $F(\mathbf{x}): \mathbb{R}^2 \rightarrow \mathbb{R}^2$ can be constructed by $F(\mathbf{x}) = (C_1(\mathbf{x}), C_2(\mathbf{x}))$, where $C_1(\mathbf{x})$ and $C_2(\mathbf{x})$ denotes the 1D image coordinates of projecting $\mathbf{x}$ onto the camera 1 and camera 2 respectively. Then, the axis-aligned grids built on the $F(\mathbf{x})$ space will exactly align with camera rays.

{\bf Proper space warping.} Based on the 2D analysis above, we define a proper space warping function as follows.
\begin{definition}
Given a region $S$ in the 3D Euclidean space and a set of cameras $\{C_i|i=1,2,...,n_c\}$ which are visible to $S$, a warping function $F:\mathbb{R}^{3}\to \mathbb{R}^{3}$ is called a {\em proper warping function}, if for any two points $\mathbf{x}_1, \mathbf{x}_2\in S$, the distance between these two points in the warp space equals to the sum of distances between these two points on all visible cameras, i.e.  $\|F(\mathbf{x}_1)-F(\mathbf{x}_2)\|_2^2=\sum_i^n \|C_i(\mathbf{x}_1)-C_i(\mathbf{x}_2)\|_2^2$.
\label{def:warp}
\end{definition}
Clearly, the warping function $F(\mathbf{x})=(C_1(\mathbf{x}), C_2(\mathbf{x}))$ in the 2D toy example is a proper 2D warping function.
Note that whether a warping function is proper or not is a local property, which only relates to the visible cameras.

\begin{figure}[!b]
  \includegraphics[width=\linewidth]{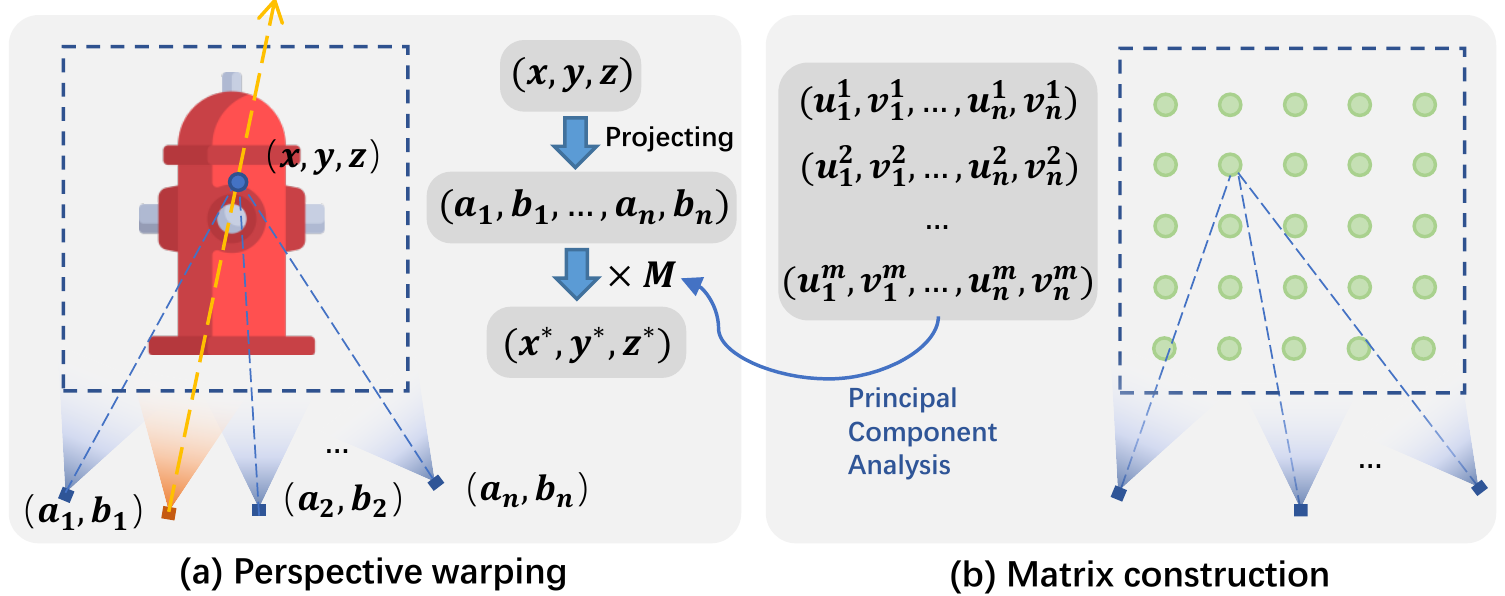} \caption{{\bf Our perspective warping method.}}
  \label{fig:method_warp}
\end{figure}

{\bf 3D perspective warping.} Given the region $S$ and the cameras $C_i$, we now would like to construct a proper 3D warping function $F(\mathbf{x}): \mathbb{R}^3 \rightarrow \mathbb{R}^3$.
We begin with an observation that the function $\mathbf{y}=G(\mathbf{x})=[C_1(\mathbf{x}),...,C_{n_c}(\mathbf{x})]: \mathbb{R}^3 \to \mathbb{R}^{2n_c}$ which maps the 3D point to its projected coordinates on all $n$ cameras $C_i$ is a proper warping function because this is a trivial construction based on the definition of a proper warping function. However, what we want to construct eventually is a function that maps a 3D space to a 3D space. Hence, we consider constructing an approximately proper warping function $F$ with the following formulation.
\begin{problem}
Let $\{\mathbf{x}_j|j=1,2,...,n_p\}$ denote $n_p$ evenly-sampled points in the local region $S$ of the original Euclidean space, we want to find a projection matrix $M\in \mathbb{R}^{3\times 2n_c}$ that maps the coordinate $\mathbf{y}_j=G(\mathbf{x}_j)\in \mathbb{R}^{2n_c}$ to $\mathbf{z}_j\in \mathbb{R}^{3}$ by $\mathbf{z}_j=M\mathbf{y}_j$, so that $M$ minimizes $\sum_j^K \|M^\intercal\mathbf{z}_j-\mathbf{y}_j\|_2^2$
\label{prob:pca}
\end{problem}
In comparison with Definition~\ref{def:warp}, Problem~\ref{prob:pca} makes two relaxations. First, we only consider the distances between sampled points $x_i\in S$ in Problem~\ref{prob:pca} while Definition~\ref{def:warp} considers arbitrary point pairs. Second, we apply a projection matrix such that the distances between $\mathbf{y}_i$ are preserved as much as possible.
It can be shown that the solution to this problem is exactly the Principle Component Analysis (PCA)~\cite{wold1987principal} on the set of projection points $\{\mathbf{y}_j\}$. The matrix $M$ is constructed from the first three eigenvectors of the covariance matrix of $\{\mathbf{y}_j\}$. Therefore, the proposed perspective warping function is $F(\mathbf{x}) = M G(\mathbf{x})$, as shown Fig.~\ref{fig:method_warp}. In our implementation, we perform a post normalization on $F(\mathbf{x})$ to make the resulting points $\{\mathbf{z}_j\}$ in the warp space located around the origin, which is introduced in details in the supplementary material.

\begin{figure}
  \includegraphics[width=\linewidth]{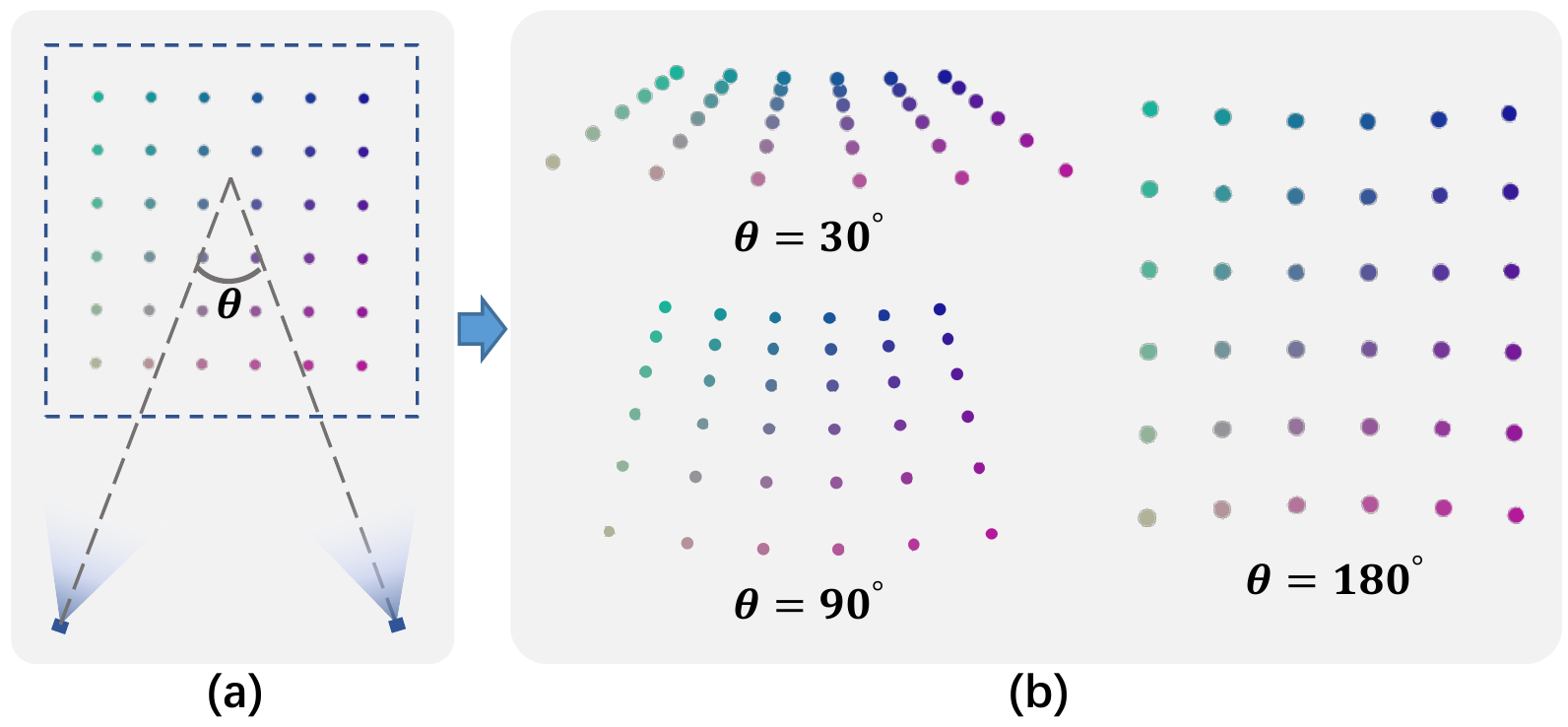} \caption{{\bf Visualization of the effect of perspective warping.} (a) Points in the original Euclidean space. (b) Points in the warp space and the corresponding camera angles.}
  \label{fig:method_warp_vis}
\end{figure}

\textbf{Intuition of $F(\mathbf{x})$}. 
$F(\mathbf{x})$ maps the region $S$ in the original space to a region around the origin of the warp space. 
Fig.~\ref{fig:method_warp_vis} shows the perspective warping with different angles between two neighboring cameras. As we can see, when the angle $\theta$ is small, the space is squashed more on the far region, which is a similar behavior to the NDC warping. In this case, the perspective warping reduces to the NDC warping if all the cameras are forward-facing. When the angle becomes larger, the warp space is more similar to the original Euclidean space.

\begin{figure}
    \centering
    \vspace{-1em}
    \includegraphics[width=\linewidth]{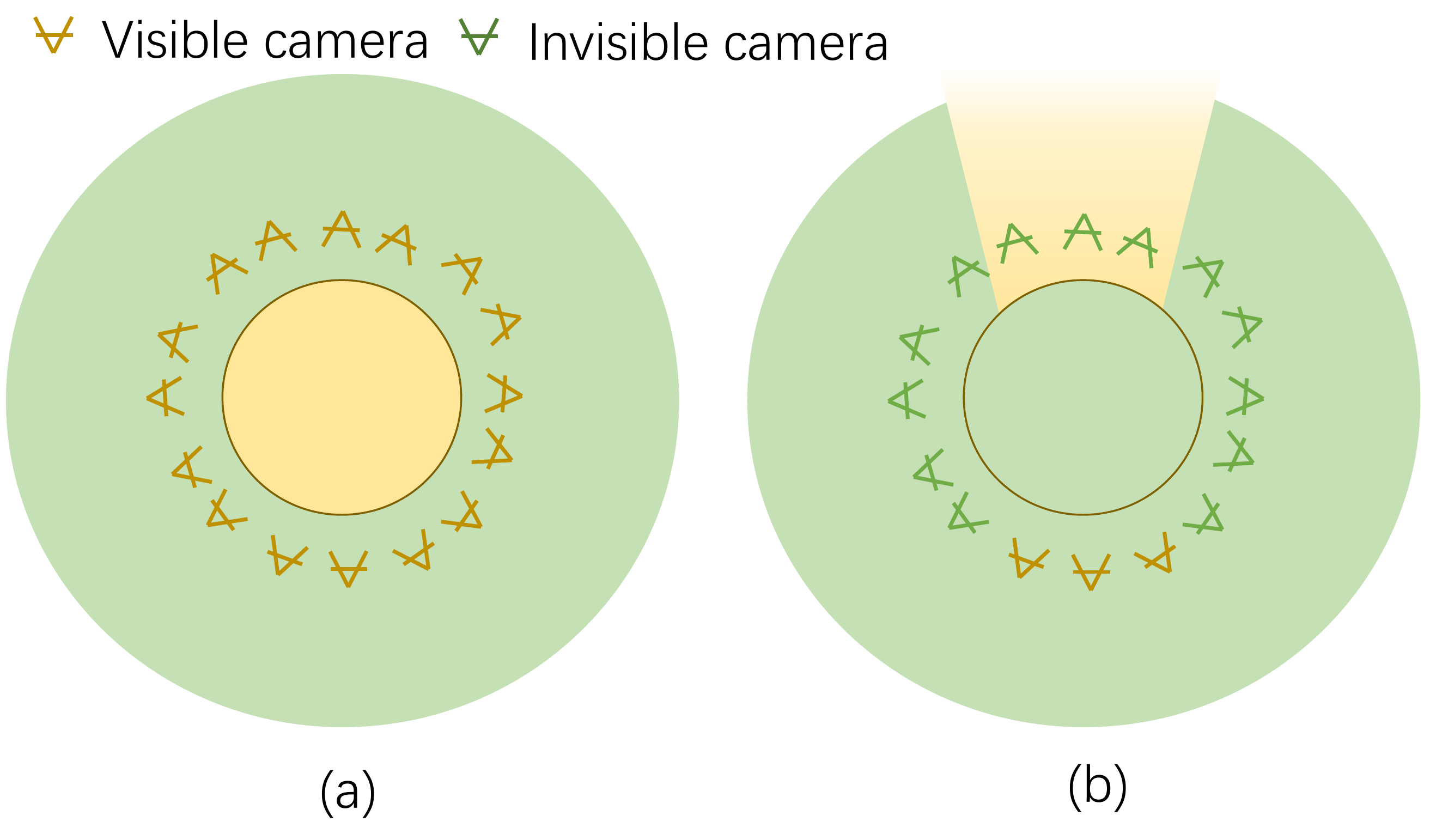}
    \caption{\textbf{Inverse sphere warping}. (a) The inner unit sphere are visible to all cameras around. Thus, the warp space is a Euclidean space, which corresponds to the 180$^{\circ}$ case of perspective warping in Fig.~\ref{fig:method_warp_vis}. (b) The outer space is only visible to a part of cameras. The outer space uses an NDC warping, which corresponds to the 30$^\circ$ case of perspective warping in Fig.~\ref{fig:method_warp_vis}.}
    \label{fig:inv_sphere}
    \vspace{-2em}
\end{figure}

\begin{figure*}[htb]
  \includegraphics[width=\linewidth]{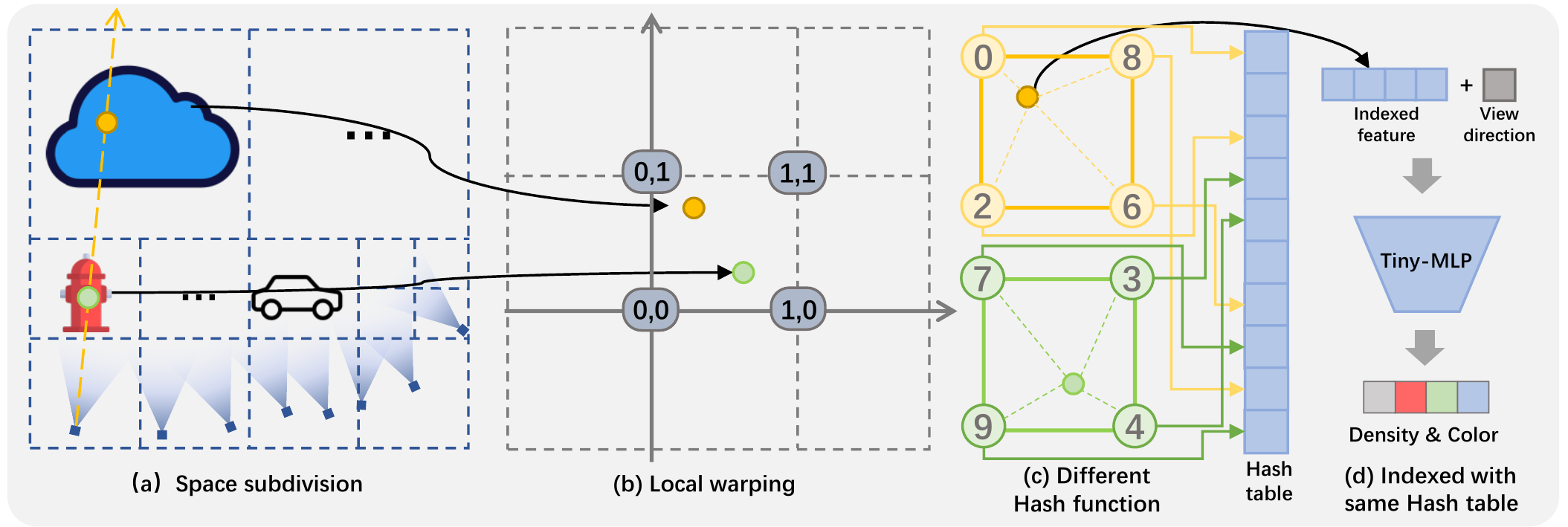} \caption{{\bf Pipeline of \ourmethod.} (a) Given a large region of interest, we subdivide the space according to the input view frustums. (b) For each sub-region, we construct a perspective warping function based on the visible cameras. The densities and colors are decoded from the scene feature vectors fetched from the same hash table (d) but using different hash functions (c).}
  \label{fig:method_overview}
  \vspace{-1em}
\end{figure*}

{\bf Relationship with inverse sphere warping}. 
We empirically show that inverse sphere warping~\cite{ZhangRSK20,BarronMVSH22} is also a handcrafted approximation of our perspective warping function.
As shown in Fig.~\ref{fig:inv_sphere}~(a), the warp space of the inverse sphere warping for the inner unit sphere is simply the original Euclidean space because all the cameras around are visible to this unit sphere, which corresponds to the $180^\circ$ case in Fig.~\ref{fig:method_warp_vis} of the perspective warping. The outer space (Fig.~\ref{fig:inv_sphere}~(b)) is only visible from a few far cameras and thus is warped similarly as the NDC warping, which corresponds to the $30^\circ$ case in Fig.~\ref{fig:method_warp_vis} of the perspective warping.

\subsection{Space subdivision}
\label{sec:sub}
In order to apply the perspective warping function $F(\mathbf{x})$, we need to specify $n_c$ cameras onto which we project the point $\mathbf{x}$. According to Definition~\ref{def:warp}, the properness of the warping function is a local property which only relates the visible cameras for the point $\mathbf{x}$. However, for a free trajectory like Fig.~\ref{fig:teaser}~(c), visible cameras for different regions are different. This motivates us to adaptively subdivide the space into different regions such that the visible cameras used in $F(\mathbf{x})$ are the same inside a region but are different across regions. In this case, in each subdivided region $S_i$, a warping function $F_i(\mathbf{x})$ is applied to map $S_i$ to the warp space. 

\textbf{Subdivision strategy}. We adopt an octree data structure to store the subdivided regions, which enables us to quickly search for a region and retrieve visible cameras. 
To construct the octree, we begin with an extremely large bounding box as the root node.
Here the box size is $512$ times the bounding box size of all camera centers, which is able to contain extremely far-away sky or other objects.
Then, starting from the octree root node, we perform a check-and-subdivide procedure. 
Specifically, on a tree node with $s$ as the side length, we retrieve all visible cameras whose view frustums intersect with this node. Then, if there is any visible camera center whose distance $d$ to the node center is $d\le \lambda s$, where $\lambda$ is preset to 3, the node is subdivided into 8 child nodes with side length of $s/2$. Otherwise, the current node is small enough and we stop subdividing it and mark it as a leaf node. For each subdivided child node, we further check the distance and repeat this procedure until we get all $n_l$ leaf nodes $\{S_i|i=1,2,...,n_l\}$ as shown in Fig.~\ref{fig:method_overview}~(a). Each leaf node is treated as the region $S$ in Problem~\ref{prob:pca}, and for those regions that are visible by more than $n_c=4$ cameras, we further select $n_c$ visible cameras by making the minimal pair-wise distance of the selected cameras as large as possible. A more detailed description of the camera selection strategy can be found in the supplementary material. By applying the warping function $F_i$, each leaf node is mapped to a region around the origin of its warp space.

\subsection{Scene representation}
\label{sec:rep}
In this section, we will introduce how to build our grid-based scene representation on the warp space, which allows color and density computation for a given point in the warp space.
Since the warping functions are different for different leaf nodes, we actually have $n_l$ different warp spaces. A naive solution would be to build $n_l$ different grid representations on each warp space. However, this would cause the number of parameters to grow with the number of leaf nodes. 
To limit parameter number, we suppose that all warping functions map different leaf nodes to the same warp space and build a hash-grid representation~\cite{mueller2022instant} on the warp space with multiple hash functions as shown in Fig.~\ref{fig:method_overview}.

\textbf{Hash grid with multiple hash functions}. Sharing the same warp space for different leaf nodes will inevitably lead to conflicts, which means two different points in two leaf nodes with different densities and colors are mapped to the same point in the warp space.
In the original Instant-NGP~\cite{mueller2022instant}, there is only one hash function to compute hash values for grid vertices.
Here, we use different hash functions for different leaf nodes to alleviate the conflict problem.
Specifically, for a point $\mathbf{x}$ in the $i$-th leaf node, we map it to $\mathbf{z}=F_i(\mathbf{x})$ in the warp space and find $z$'s eight neighboring grid vertices $\hat{\mathbf{v}}$ with integer coordinates.
Then, we compute a hash value for each vertex $\hat{\mathbf{v}}$ by a hash function conditioned on the leaf node index $i$ as follows
\begin{equation}
{\rm Hash}_i(\hat{\mathbf{v}}) = \left(\bigoplus_{k=1}^{3}\hat{\mathbf{v}}_k\pi_{i,k} + \Delta_{i,k} \right)\mod L,
\label{eq:subhash}
\end{equation}
where $\bigoplus$ denotes the bitwise xor operation, and both $\left\{\pi_{i,k}\right\}$ and $\left\{\Delta_{i,k}\right\}$ are random large prime numbers, which are fixed for a specific leaf node, $k=1,2,3$ means the index of $x,y,z$ coordinate of the warp space, $L$ is the length of the hash table. The computed hash value will be used in indexing the hash table to retrieve a feature vector for the vertex $\hat{\mathbf{v}}$ and then the feature vector of the point $z$ is trilinearly-interpolated from 8 vertex feature vectors. Finally, the feature vector of $z$ and the view direction $\mathbf{d}$ are fed into a tiny MLP network to produce color and density for the point $z$, as shown in Fig.~\ref{fig:method_overview}~(d).

\textbf{Intuition of Eq.~\ref{eq:subhash}}. 
Using different numbers $\pi_{i,k}$ and different offsets $\Delta_{i,k}$ leads to different hash functions in Eq.~\ref{eq:subhash}. 
We use an example in Fig.~\ref{fig:method_overview} to show how Eq.~\ref{eq:subhash} works: The green point and the yellow point in two different leaf nodes (Fig.~\ref{fig:method_overview}~(a)) are mapped by two different warping functions to the same warp space (Fig.~\ref{fig:method_overview}~(b)). Suppose that both points reside in the same voxel of the warp space.  Although the coordinates of their neighboring vertices are the same, the two points will use different hash functions Eq.~\ref{eq:subhash} to compute different hash values for these neighboring vertices (Fig.~\ref{fig:method_overview}~(c)). 
Then, the computed hash values will be used in indexing the same hash table to retrieve feature vectors (Fig.~\ref{fig:method_overview}~(d)). In this case, two points from different leaf nodes share the same neighboring vertices in the warp space but retrieve different vertex feature vectors, which greatly reduces the probability of conflicts.
Though some conflicts still remain, they can naturally be resolved by the tiny MLP during the optimization process as observed by Instant-NGP~\cite{mueller2022instant}.

\subsection{Perspective sampling} 
\label{sec:method_sampling}
A proper warping function $F$ also gives us a guideline for sampling points on rays in volume rendering. 
According to Definition~\ref{def:warp}, the distance between two points in the proper warp space equals the sum of distances between two projected points on image planes.
In this case, by uniformly sampling the points in the warp space, we get a non-uniform sampling in the original Euclidean space but an approximately uniform sampling on images, which improves the sampling efficiency and brings more stable convergence. 
Specifically, considering a sample point $\mathbf{x}_i = \mathbf{o} + t_i\mathbf{d}$ where $\mathbf{o},\mathbf{d}$ are the camera origin and direction respectively, we first compute the Jacobian matrix $J_i \in \mathbb{R}^{3\times 3}$ of the perspective warping function $F$ at $\mathbf{x}_i$. 
Then, the next sample point $\mathbf{x}_{i+1} = \mathbf{x}_{i} + \frac{l}{\|J_i\mathbf{d}\|_2} \mathbf{d}$, where $l$ is a preset parameter controlling the sampling interval and we make a linear approximation here as discussed in the supplementary material.

\subsection{Rendering with perspective warping}
Based on the descriptions above, we now summarize the rendering procedure in two stages: the preparation stage and the actual rendering stage. (1) In the preparation stage, we subdivide the original space according to the view frustums of the cameras (Sec.~\ref{sec:sub}), and construct the local warping functions based on the selected cameras for each sub-region (Sec.~\ref{sec:per}). (2) In the actual rendering stage, we follow the framework of volume rendering to render a pixel color, by sampling points on the camera ray (Sec.~\ref{sec:method_sampling}) and conducting weighted accumulation of the sampled colors (Sec.~\ref{sec:pre}). The sampled densities and colors are fetched from the multi-resolution hash grid (Sec.~\ref{sec:rep}).

\subsection{Training}
The training loss is defined as
\begin{equation}
    \mathcal{L} = \mathcal{L}_{recon(c(r), c_{{\rm gt}})} + \lambda_{\rm Disp}\mathcal{L}_{\rm Disp} + \lambda_{\rm TV}\mathcal{L}_{\rm TV},
\end{equation} 
where $\mathcal{L}_{recon(c(r), c_{\rm gt})}=\sqrt{(c(r) - c_{{\rm gt}})^2 + \epsilon}$ is a color reconstruction loss~\cite{BarronMVSH22} with $\epsilon = 10^{-4}$.
$\mathcal{L}_{\rm Disp}$ and $\mathcal{L}_{\rm TV}$ are two regularization losses. The first is a disparity loss $\mathcal{L}_{\rm Disp}$ that encourages the disparity (inverse depth) not excessively large, which is useful to reduce the floating artifacts. 
The second is a total variance loss~\cite{rudin1994total} $L_{\rm TV}$ that encourages the points at the borders of two neighboring octree nodes $i,j$ to have similar densities and colors. For all losses, we provide more details in the supplementary material. 

\section{Experiments}

\begin{figure}[!t]
  \includegraphics[width=\linewidth]
  {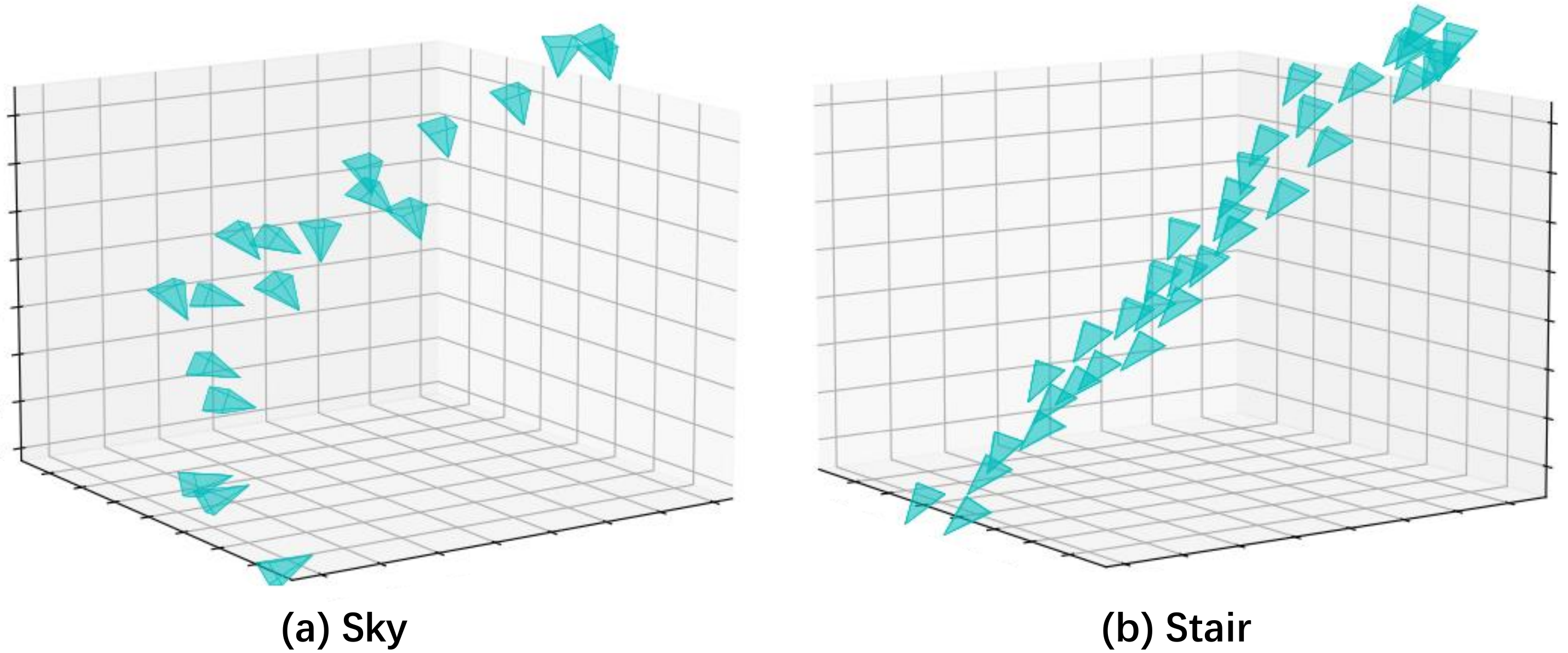} \caption{{\bf Trajectories of ``Sky" and ``Stair" in the Free dataset.}}
  \label{fig:traj}
  \vspace{-1em}
\end{figure}

\begin{figure*}[!b]
  \vspace{-1em}
  \includegraphics[width=\linewidth]{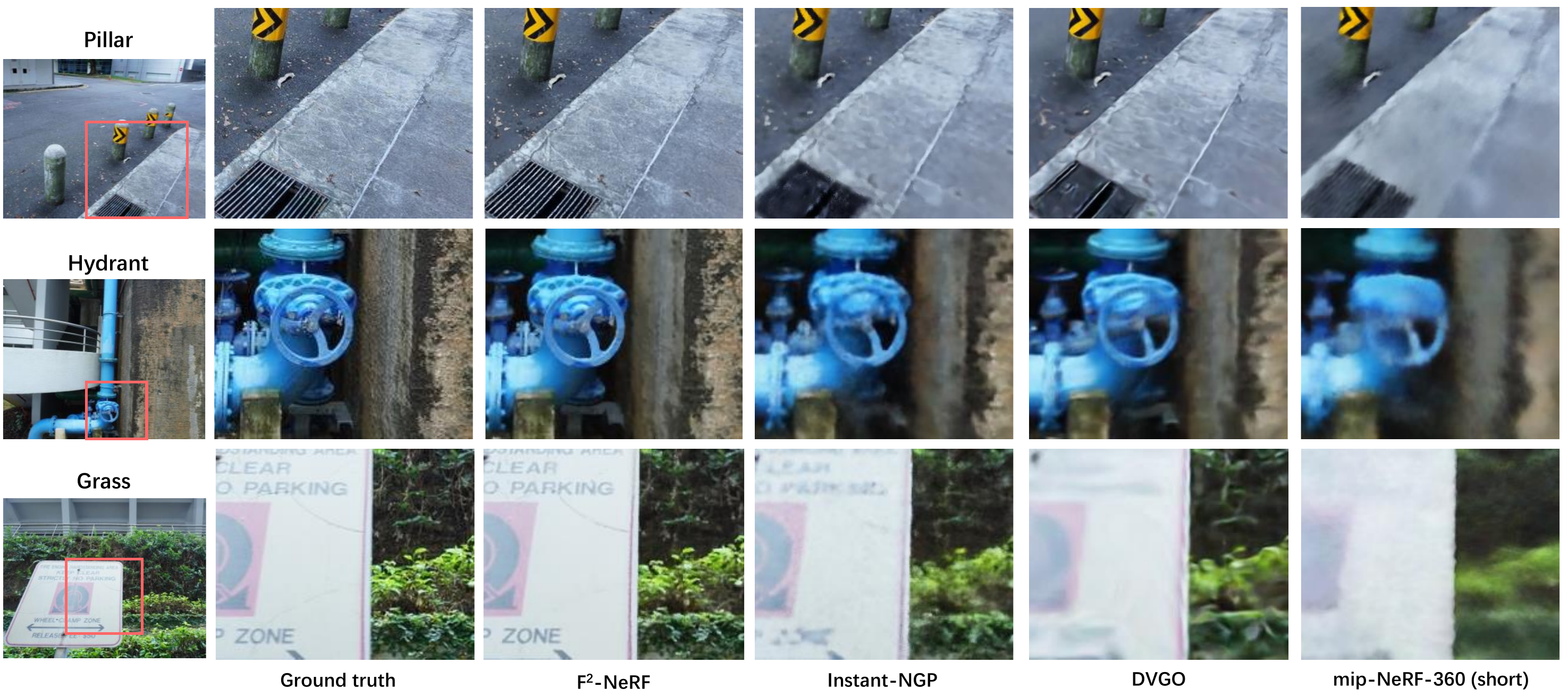} \caption{{\bf Visual comparison on the Free dataset.}}
  \label{fig:compare_free}
\end{figure*}

\subsection{Experimental Settings}
{\bf Datasets.} We use three datasets for our evaluation. (1) A new unbounded dataset with free trajectories that we collected (called the {\em Free dataset}). The Free dataset contains seven scenes. Each scene has a narrow and long input camera trajectory and multiple focused foreground objects, which is extremely challenging to build a neural representation for the NVS task. Two of these trajectories are shown in Fig.~\ref{fig:traj}; (2) LLFF dataset~\cite{MildenhallSCKRN19}, which contains eight real unbounded forward-facing scenes with complex geometries; and (3) NeRF-360-V2 dataset~\cite{BarronMVSH22}, which contains seven unbounded 360-degree inward-facing outdoor and indoor scenes. For all the datasets, we follow the commonly-adopted settings that set one of every eight images as testing images and the other as the training set. We use three metrics, PSNR, SSIM, and LPIPS$_{\rm VGG}$, for evaluation.

{\bf Baselines.} We compare \ourmethod with the state-of-the-art fast NeRF training methods, the voxel-based methods (1) DVGO~\cite{SunSC22}, (2) Plenoxels~\cite{YuFTCR22} and the hash-grid based method (3) Instant-NGP~\cite{mueller2022instant}. We also report the results of MLP-based NeRF methods including NeRF++~\cite{ZhangRSK20}, mip-NeRF~\cite{BarronMTHMS21}, and mip-NeRF-360~\cite{BarronMVSH22}.
Note that both \ourmethod and Instant-NGP~\cite{mueller2022instant} are trained for 20k steps with the same batch size but we adopt a LibTorch~\cite{paszke2019pytorch} implementation while Instant-NGP uses a CUDA implementation, which is faster (6min) than ours (13min). All training times are evaluated on a single 2080Ti GPU.
Implementation details of \ourmethod are included in the supplementary material.

{\bf Warping functions.} Different warping functions are adopted for baseline methods on different datasets. On the LLFF dataset, all baselines use the NDC warping function, and on both the Free dataset and the NeRF-360-V2 dataset, all baseline methods use the inverse sphere warping function, except Instant-NGP. In the Instant-NGP, we follow the official implementation to enlarge the ray marching bounding box to represent backgrounds and carefully tune the scale parameters on different scenes to achieve the best performance. In comparison, \ourmethod always uses the perspective warping for all datasets.

\begin{table}[t!]
    \centering
    \begin{tabular}{@{}l@{\hskip 3pt}|@{\hskip 3pt}c@{\hskip 8pt}c@{\hskip 8pt}c@{\hskip 8pt}c@{}}
    \hline
    Method & Tr. time & PSNR{\scriptsize$\uparrow$} & SSIM{\scriptsize$\uparrow$} & LPIPS{\scriptsize(VGG)$\downarrow$} \\
    \hline\hline
    {\small NeRF++~\cite{ZhangRSK20}} & hours & 23.47 & 0.603 & 0.499 \\
    {\small mip-NeRF-360~\cite{BarronMVSH22}} & hours & {\bf 27.01} & {\bf 0.766} & {\bf 0.295} \\
    \hline
    {\small mip-NeRF-360$_{\rm short}$} & 30m & 22.04 & 0.537 & 0.586 \\
    {\small Plenoxels~\cite{YuFTCR22}} & 25m & 19.13 & 0.507 & 0.543 \\
    {\small DVGO~\cite{SunSC22}} & 21m & 23.90 & 0.651 & 0.455 \\
    {\small Instant-NGP~\cite{mueller2022instant}} & 6m & 24.43 & 0.677 & 0.413 \\
    {\small \ourmethod} & 12m & {\bf 26.32} & {\bf 0.779} & {\bf 0.276} \\
    \hline
    \end{tabular}
    \vspace{-1em}
    \caption[]{{\bf Results on the Free dataset}. In mip-NeRF-360$_{\rm short}$, we early stop the training to make them finished in 30 minutes. Training times are evaluated on a 2080ti GPU.
    }
    \label{tab:compare_free}
    \vspace{-1em}
\end{table}

\subsection{Comparative studies}
We report the quantitative comparisons on the Free dataset in Table~\ref{tab:compare_free}. \ourmethod achieves the best rendering quality among all the fast-training NeRFs. The results for qualitative comparison are shown in Fig.~\ref{fig:compare_free}. The synthesized images of DVGO~\cite{SunSC22} and Plenoxels~\cite{YuLTLNK2021} are blurred due to their limited resolutions to represent such a long trajectory.
The results of Instant-NGP look sharper but are not clear enough due to its unbalanced scene space organization. In comparison, \ourmethod takes advantage of the perspective warping and the adaptive space subdivision to fully exploit the representation capacity, which enables \ourmethod to produce better rendering quality. Meanwhile, we find that training a mip-NeRF-360 on the Free dataset for a long time is also able to render clear images. The reason is that during the training process, the large MLP networks used by mip-NeRF-360 are able to gradually concentrate on foreground objects and adaptively allocate more capacity to these foreground objects. However, these MLP networks have to spend a long training time for convergence on the Free Trajectory dataset. With a short training time (30 minutes), the results of mip-NeRF-360 contain many foggy artifacts.

We also evaluate our method on the widely-used unbounded forward-facing dataset (LLFF) and 360$^\circ$ object-centric dataset (NeRF-360-V2) to show the compatibility of the perspective warping with these two kinds of specialized camera trajectories. On both datasets, \ourmethod achieves comparable results to the other fast NeRF methods. Note these baseline fast NeRF methods adopt the specially-designed NDC warping or inverse sphere warping for the LLFF dataset or the NeRF-360-V2 dataset while \ourmethod always uses the same perspective warping for all datasets. This demonstrates the compatibility of the perspective warping with different trajectories.

\begin{table}[t!]
    \centering
    \begin{tabular}{@{}l@{\hskip 3pt}|@{\hskip 3pt}c@{\hskip 8pt}c@{\hskip 8pt}c@{\hskip 8pt}c@{}}
    \hline
    Method & Tr. time & PSNR{\scriptsize$\uparrow$} & SSIM{\scriptsize$\uparrow$} & LPIPS{\scriptsize(VGG)$\downarrow$} \\
    \hline\hline
    {\small NeRF++~\cite{ZhangRSK20}} & hours & 26.21 & 0.729 & 0.348\\
    {\small mip-NeRF-360~\cite{BarronMTHMS21}} & hours & {\bf 28.94} & {\bf 0.837} & {\bf 0.208} \\
    \hline
    {\small Plenoxels~\cite{YuFTCR22}} & 22m & 23.35 & 0.651 & 0.471 \\
    {\small DVGO~\cite{SunSC22}} & 16m & 25.42 & 0.695 & 0.429 \\
    {\small Instant-NGP~\cite{mueller2022instant}} & 6m  & 26.24 & 0.716 & 0.404 \\
    {\small \ourmethod} & 14m & {\bf 26.39} & {\bf 0.746} & {\bf 0.361} \\
    \hline
    \end{tabular}
    \vspace{-1em}
    \caption[]{{\bf Results on the NeRF-360-V2 dataset.}}
    \label{tab:compare_360}
    \vspace{-0em}
\end{table}

\begin{table}[t!]
    \centering
    \begin{tabular}{@{}l@{\hskip 3pt}|@{\hskip 3pt}c@{\hskip 8pt}c@{\hskip 8pt}c@{\hskip 8pt}c@{}}
    \hline
    Method & Tr. time & PSNR{\scriptsize$\uparrow$} & SSIM{\scriptsize$\uparrow$} & LPIPS{\scriptsize(VGG)$\downarrow$} \\
    \hline\hline
    {\small NeRF~\cite{MildenhallSTBRN20}} & hours & 26.50 & 0.811 & 0.250 \\
    {\small mip-NeRF~\cite{BarronMTHMS21}}& hours & {\bf 26.60} & {\bf 0.814} & {\bf 0.246} \\
    \hline
    {\small Plenoxels~\cite{YuFTCR22}} & 17m & 26.29 & 0.839 & 0.210 \\
    {\small DVGO~\cite{SunSC22}} & 11m & 26.34 & 0.838 & 0.197 \\
    {\small TensoRF~\cite{ChenXGYS22}} & 48m & {\bf 26.73} & 0.839 & 0.204 \\ 
    {\small Instant-NGP~\cite{mueller2022instant}} & 6m  & 25.09 & 0.758 & 0.267 \\
    {\small \ourmethod} & 13m & 26.54 & {\bf 0.844} & {\bf 0.189} \\
    \hline
    \end{tabular}
    \vspace{-1em}
    \caption[]{{\bf Results on the LLFF dataset.}}
    \label{tab:compare_llff}
    \vspace{-2em}
\end{table}

\subsection{Ablation studies}
\begin{figure}[!t]
  \includegraphics[width=\linewidth]{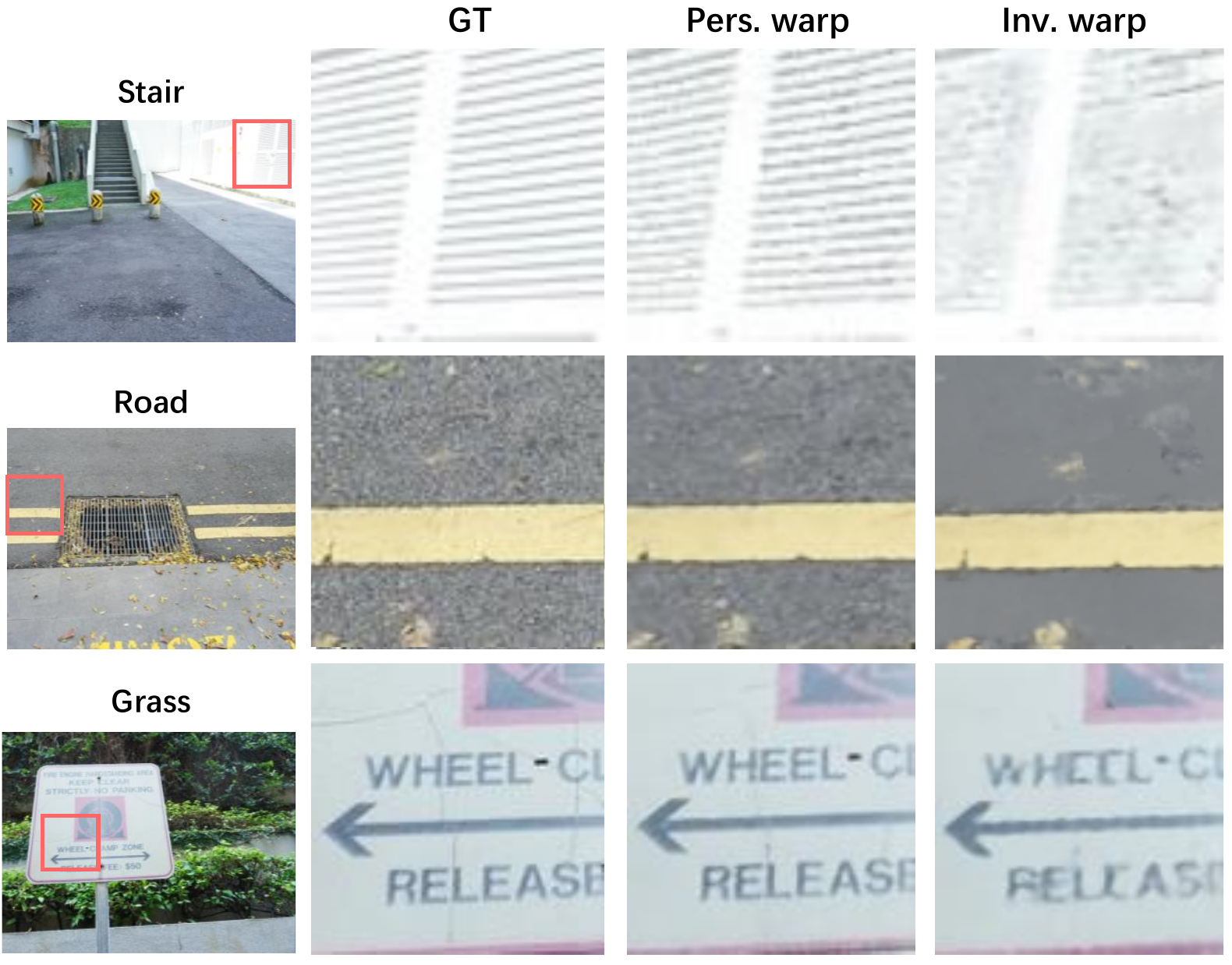} \caption{{\bf Visual comparison of different warping techniques.} Our perspective warping renders more clear images than inverse sphere warping~\cite{BarronMVSH22}. }
  \label{fig:ablation}
\end{figure}

We conduct ablation studies on the ``pillar" from the Free dataset. In the ablation studies, we use the multi-resolution hash grid~\cite{mueller2022instant} as the scene representation and change the warping functions and the sampling strategies. 
The warping functions include the inverse sphere warping (Inv. warp), the perspective warping (Pers. warp), and no warping (w/o warp). In the implementation of inverse sphere warping on the Free dataset, we use the bounding sphere of all camera positions as the foreground inner sphere and treat the space outside the sphere as backgrounds. For the point sampling strategies, we consider sampling by disparity (inverse-depth) (Disp. Sampling) used in mip-NeRF-360~\cite{BarronMVSH22}, sampling by the exponential function (Exp. Sampling) used in Instant-NGP~\cite{mueller2022instant}, and our perspective sampling (Sec.~\ref{sec:method_sampling}). 
The quantitative results are shown in Table~\ref{tab:ablation} and some qualitative results are shown in Fig.~\ref{fig:ablation}.

As shown in Table~\ref{tab:ablation}, the basic model (A) without space warping and using disparity sampling performs worst. The model (B) with Inv. warping improves the performances, which shows better compatibility with unbounded scenes compared with no warping (A). The model (C) replaces the disparity sampling with the exponential sampling, which makes the results better. The model (D) uses the proposed perspective warping, whose performance increases drastically compared to the inverse sphere warping. This demonstrates the effectiveness of our perspective warping on free trajectories. The model (E) further applies perspective sampling, which produces the best performance.

\begin{table}[!t]
    \centering
    \begin{tabular}{@{}l@{\hskip 3pt}|@{\hskip 3pt}c@{\hskip 8pt}c@{\hskip 8pt}c@{}}
    \hline
    Setting & PSNR{\scriptsize$\uparrow$} & SSIM{\scriptsize$\uparrow$} & LPIPS{\scriptsize$\downarrow$} \\
    \hline
    \hline
    A) w/o warp + Disp. sample & 26.81 & 0.693 & 0.385 \\
    B) Inv. warp + Disp. sample & 27.09 & 0.711 & 0.358 \\
    C) Inv. warp + Exp. sample & 27.86 & 0.751 & 0.289 \\
    D) Pers. warp + Exp. sample & 28.65 & 0.796 & 0.221 \\
    E) Pers. warp + Pers. sample & {\bf 28.76} & {\bf 0.798} & {\bf 0.219}  \\
    \hline
    \end{tabular}
    \vspace{-1em}
    \caption[]{{\bf Ablation study on the ``pillar" scene of the Free dataset.} \ourmethod with perspective warping and sampling achieves the best quantitative results.}
    \label{tab:ablation}
    \vspace{-1.5em}
\end{table}

\section{Conclusion}
In the NVS task of unbounded scenes, previous NeRF methods mainly rely on the NDC warping or the inverse sphere warping to process the forward-facing camera trajectory or the 360$^\circ$ object-centric trajectory. 
In this paper, we conduct an in-depth analysis of the space warping function and propose a novel perspective warping, which is able to handle arbitrary input camera trajectories. Based on the perspective warping, we develop a novel \ourmethod in the grid-based NeRF framework. Extensive experiments demonstrate that the proposed \ourmethod is able to render high-quality images with arbitrary trajectories while only requiring a few minutes of training time.

\noindent {\bf Potential negative societal impact.} 
\ourmethod can be possibly used for misleading fake image generation.

\noindent {\bf Acknowledgement.} This study is supported by the Ministry of Education, Singapore, under its MOE AcRF Tier 2 (MOE-T2EP20221-0012), NTU NAP, and under the RIE2020 Industry Alignment Fund – Industry Collaboration Projects (IAF-ICP) Funding Initiative, as well as cash and in-kind contribution from the industry partner(s).
Lingjie Liu and Christian Theobalt have been supported by the ERC Consolidator Grant 4DReply (770784).

{\small
\bibliographystyle{ieee_fullname}
\bibliography{egbib}
}

\clearpage

\setcounter{section}{0}
\renewcommand{\thesection}{\Alph{section}} 
\begin{center}
    {\bf\LARGE - Supplementary -}
\end{center}

\section{Additional Implementation Details}
\begin{figure}[htb]
  \includegraphics[width=\linewidth]{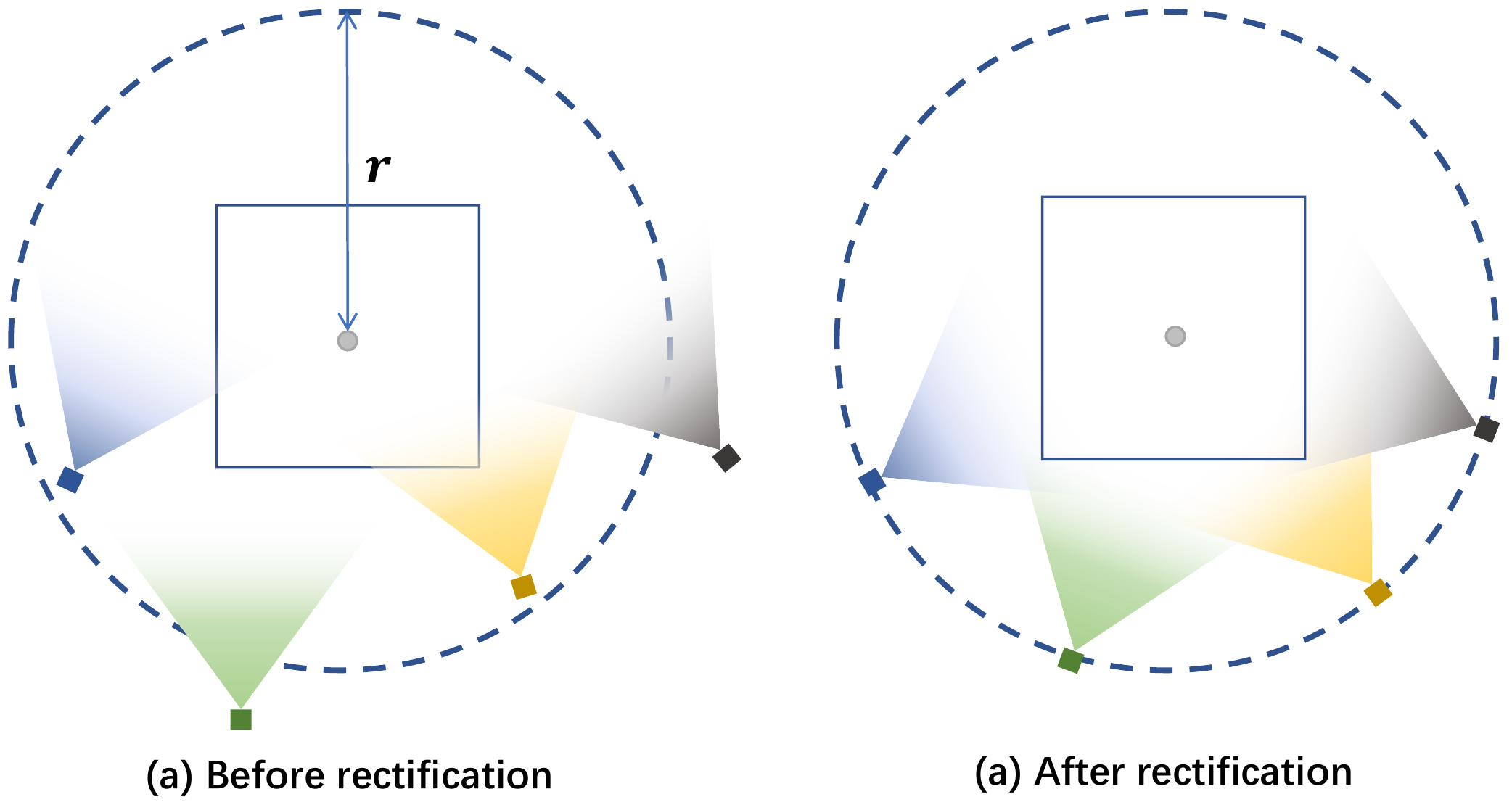} \caption{{\bf Camera rectification.}}
  \label{fig:camera_rectification}
\end{figure}

\subsection{Camera processing}
{\bf Camera rectification.} After the procedure of space-subdivision (Sec~3.3 of the main paper), for each sub-region $S$, we obtain a set of visible cameras whose view frustums intersect with $S$. 
We note that visible cameras are not suitable to be directly used in computing of the perspective warping, because some cameras do not fully cover the region and look at the region as shown in Fig.~\ref{fig:camera_rectification}(a).
Thus, we propose an empirical but effective camera rectification strategy, that we rotate the camera view directions to make them look at the center of the region $S$. This simple strategy helps ensure most points inside $S$ can be warped to meaningful coordinates. Moreover, we find that aligning their distances to the center of $S$ with the same distance $r$ can help improve the rendering quality, as shown in Fig.~\ref{fig:dis_alignment}. Here $r$ is empirically set as the mean distance to the region center among the 1/4 nearest visible cameras.

{\bf Camera selection.}
When the number of visible cameras is larger than $n_c = 4$, we select a subset of the visible cameras for better computational efficiency. We select the cameras based on the farthest point sampling: First, we randomly select a camera as the seed, and then we repeatedly add the farthest visible camera for $n_c - 1$ times.

\begin{figure}[htb]
  \includegraphics[width=\linewidth]{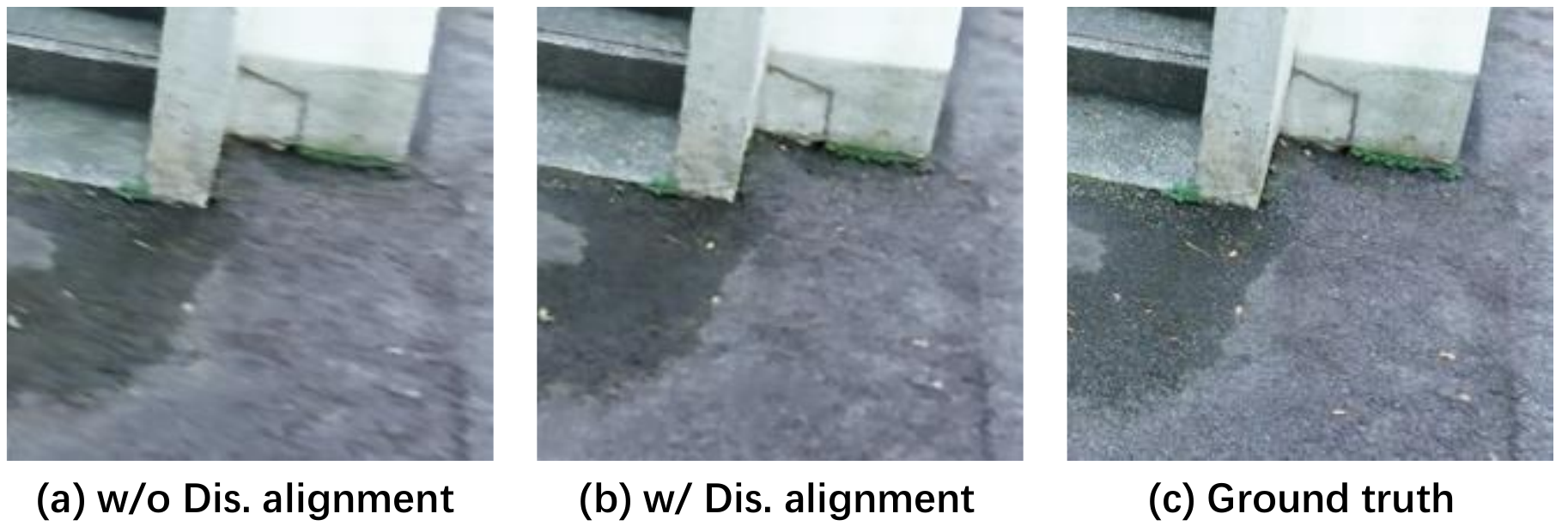} \caption{{\bf The effect of distance alignment.}}
  \label{fig:dis_alignment}
\end{figure}

\subsection{Construction of $M$}
In this section, we give a detailed description of how we construct the matrix $M$ for our perspective warping.

{\bf Principal component analysis.}
Given the region $S$ with the selected cameras, we first uniformly sample $n = 32^3$ points $\{ {\bf x}_i\}$ inside $S$. Then, we project the points to the selected cameras, concatenate the projected coordinates, and obtain the high-dimensional coordinates $\left\{\left[C_1({\bf x}_i), ..., C_{n_c}({\bf x}_i) \right] = \left[u_1^i, v_1^i, ..., u_{n_c}^i, v_{n_c}^i \right]\right\}.$ These coordinates are formed in a coordinate matrix $K \in \mathbb{R}^{2n_c \times n}$, then we compute the covariance matrix $Q = (K - \overline{K})(K - \overline{K})^{\top}$, where $\overline{K}$ is the mean coordinate of all projected coordinates. By eigendecomposition, we obtain the matrix $M' \in \mathbb{R}^{3\times 2n_c}$ formed by the eigenvectors with the first three largest eigenvalues. The matrix $M'$ defines the directions of the projection axes.

\begin{figure}[t]
  \includegraphics[width=\linewidth]{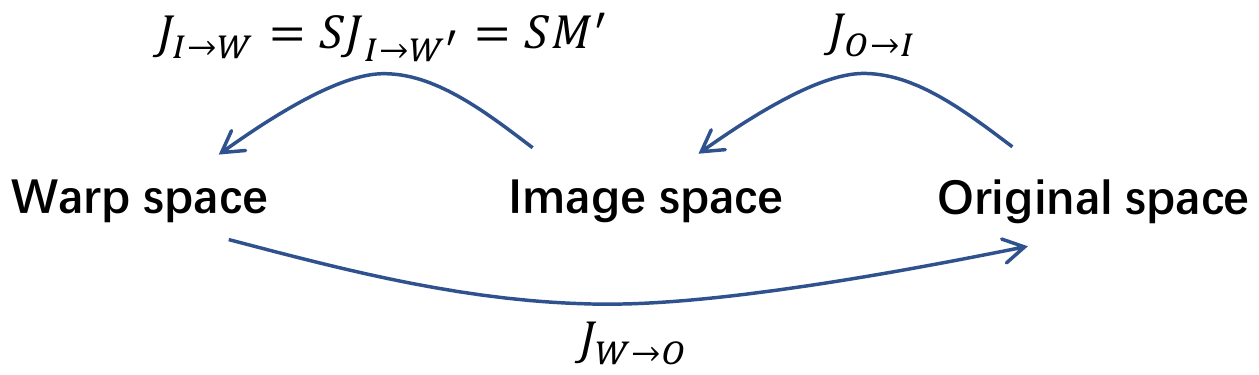} \caption{{\bf The Jacobian matrices among different spaces.}}
  \label{fig:dis_alignment}
\end{figure}

{\bf Computing the axis length.}
After the matrix $M' \in \mathbb{R}^{3\times 2n_c}$ is found, we now need to perform a post normalization by scaling each axis. Specifically, we would like to find three proper scale parameters $\{s_1, s_2, s_3\}$, and $M = SM'$, where $S \in \mathbb{R}^{3\time 3}$ is the diagonal matrix formed by $\{s_1, s_2, s_3\}$. The key idea of these scaling parameters is that we expect the unit length in the warp space can be approximately aligned with the unit length in the image space. More specifically, for each axis in the warp space, when a point moves along the axis by a unit length, we expect the maximum spatial transition of all image coordinates to be approximately a pixel length. 

We take a point ${\bf x}$ inside the region $S$ for example. 
Let us denote the Jacobian matrix from the original space to the image space as $J_{O\rightarrow I} \in \mathbb{R}^{3\times 2n_c} $ derived from the image projection function, and the Jacobian matrix from the image space to the warp space as $J_{I\rightarrow W}=M=SM'$. 
Our target is to compute the Jacobian matrix $J_{W\rightarrow I} \in \mathbb{R}^{2n_c \times 3}$ and put a constraint that the maximum value of each column vector of $J_{W\rightarrow I}$ equals one.
Note $J_{W\rightarrow I}$ may not be directly computed by inverting $J_{I\rightarrow W}$, which is not a square matrix. Alternatively, we present it by $J_{W\rightarrow I} = J_{O\rightarrow I}J_{W\rightarrow O}$, which can be further represented by 
\begin{equation}
\begin{split}
J_{W\rightarrow I} &= J_{O\rightarrow I}J_{O\rightarrow W}^{-1} \\
&=J_{O\rightarrow I}(J_{I \rightarrow W}J_{O\rightarrow I})^{-1} \\
&=J_{O\rightarrow I}(SM'J_{O\rightarrow I})^{-1} \\
&=J_{O\rightarrow I}(M'J_{O\rightarrow I})^{-1}S^{-1}.
\end{split}
\end{equation}

What we expect is that the maximum value of each column vector of $J_{W\rightarrow I}$ equals one. This constraint can solve the values of $\{s_1, s_2, s_3\}$ for the example point ${\bf x}$. For all the sampled points $\{{\bf x}_i\}$, we take the average values of $\{s_1, s_2, s_3\}$ for our final scale parameters.

\subsection{Perspective sampling}
As stated in the main paper (Sec 3.5), when sampling points in ray marching, we perform uniform sampling on the warp space, and we get a non-uniform sampling in the original space. To be specific, for the current sample point ${\bf x}_i = {\bf o} + t_i{\bf d}$, we expect to find the next sample point ${\bf x}_{i+1} = {\bf x}_i + \delta_i{\bf d}$, such that $\|F({\bf x}_{i+1}) - F({\bf x}_{i})\|_2 = l$. Here $l$ is the parameter controlling sample density and we empirically set $l=\sqrt{3}$, i.e., the diagonal length of the unit cube in the warp space~\cite{mueller2022instant}. To compute the marching step $\delta_i$ in the original space efficiently, we perform a linear approximation that
\begin{equation}
    F({\bf x}_{i+1}) \approx F({\bf x}_{i}) + \delta_i \cdot J_i{\bf d},
\end{equation}
Where $J_i$ is the Jacobian matrix at ${\bf x}_i$ from the original space to the warp space. Hence, the distance between $F({\bf x}_{i+1})$ and $F({\bf x}_i)$ is approximated by $\delta_i \|J_i{\bf d}\|_2$. We let it equals $l$ and get $\delta_i = \frac{l}{\|J_i{\bf d}\|_2}$. 

\subsection{Loss functions}
As described in the main paper, the loss of training is defined as
\begin{equation}
    \mathcal{L} = \mathcal{L}_{recon(c(r), c_{{\rm gt}})} + \lambda_{\rm Disp}\mathcal{L}_{\rm Disp} + \lambda_{\rm TV}\mathcal{L}_{\rm TV},
\end{equation} 
where the first term $\mathcal{L}_{recon(c(r), c_{\rm gt})}=\sqrt{(c(r) - c_{{\rm gt}})^2 + \epsilon}$ is a color reconstruction loss~\cite{BarronMVSH22} with $\epsilon = 10^{-4}$, and the last two terms are the regularization losses.

The disparity loss $\mathcal{L}_{\rm Disp}$ of the sampled rays is defined by
\begin{equation}
\mathcal{L}_{\rm Disp} = \frac{1}{n_r}\sum_{k}{\rm disp}_k^2,
\end{equation}
where the disparity of a ray is computed by the weighted sum of the sampled inverse distance that ${\rm disp} = \sum_i w_i\frac{1}{t_i}$, and $\{w_i\}$ are the weights computed by volume rendering.

The aim of total variation loss $\mathcal{L}_{\rm TV}$ is to encourage the border points of two neighboring octree nodes to have similar densities and colors. To achieve this goal, in each training iteration, we randomly sample $n_b=8192$ points on the borders of the octree nodes, then the loss is defined by
\begin{equation}
    \mathcal{L}_{\rm TV} = \frac{1}{n_b} \sum_k \|{\rm feat}_0^k - {\rm feat}_1^k\|_2^2.
\end{equation}
Here, for each sample point $k$, ${\rm feat}_0^k$ and ${\rm feat}_1^k$ are the feature vectors fetched from the hash table using two different functions conditioned on its two neighboring octree nodes.

In LLFF dataset we set $\lambda_{\rm Disp}=2.5\times 10^{-4}, \lambda_{\rm TV}=10^{-1}$, and in Free dataset and NeRF-360-V2 dataset we set $\lambda_{\rm Disp}=10^{-3}, \lambda_{\rm TV}=10^{-1}$.

\subsection{More implementation details}
{\bf Architecture details.} We follow a similar setting to Instant-NGP~\cite{mueller2022instant} and use the hash table with $16$ levels, and each level contain $2^{19}$ feature vectors with dimension of $2$. The fetched hash feature vectors of size 32 are fed to a tiny MLP with one hidden layer of width 64, to get the scene features and the volume densities, then, the scene features are concatenated with the spherical harmonics encoding of view directions and are fed to another rendering MLP with two hidden layers of width 64 to get the RGB colors.

{\bf Training details.} We follow Instant-NGP~\cite{mueller2022instant} and set the fixed batch size of point samples as 256k while the batch size of rays is dynamic, depending on the average sampled points on rays. We train the parameters with Adam optimizer~\cite{kingma2014adam}, whose learning rate linearly grows from zero to $1\times 10^{-1}$ in the first 1k steps and the decay to $10^{-2}$ at the end of training with cosine scheduling. For all the scenes in the experiments, we train \ourmethod for 20k steps. We implement \ourmethod using LibTorch~\cite{paszke2019pytorch}. The training time depends on the scene's complexity, and for most cases, it is between 10 minutes and 15 minutes on a single Nvidia 2080Ti GPU.

\begin{figure}[htb]
  \includegraphics[width=\linewidth]{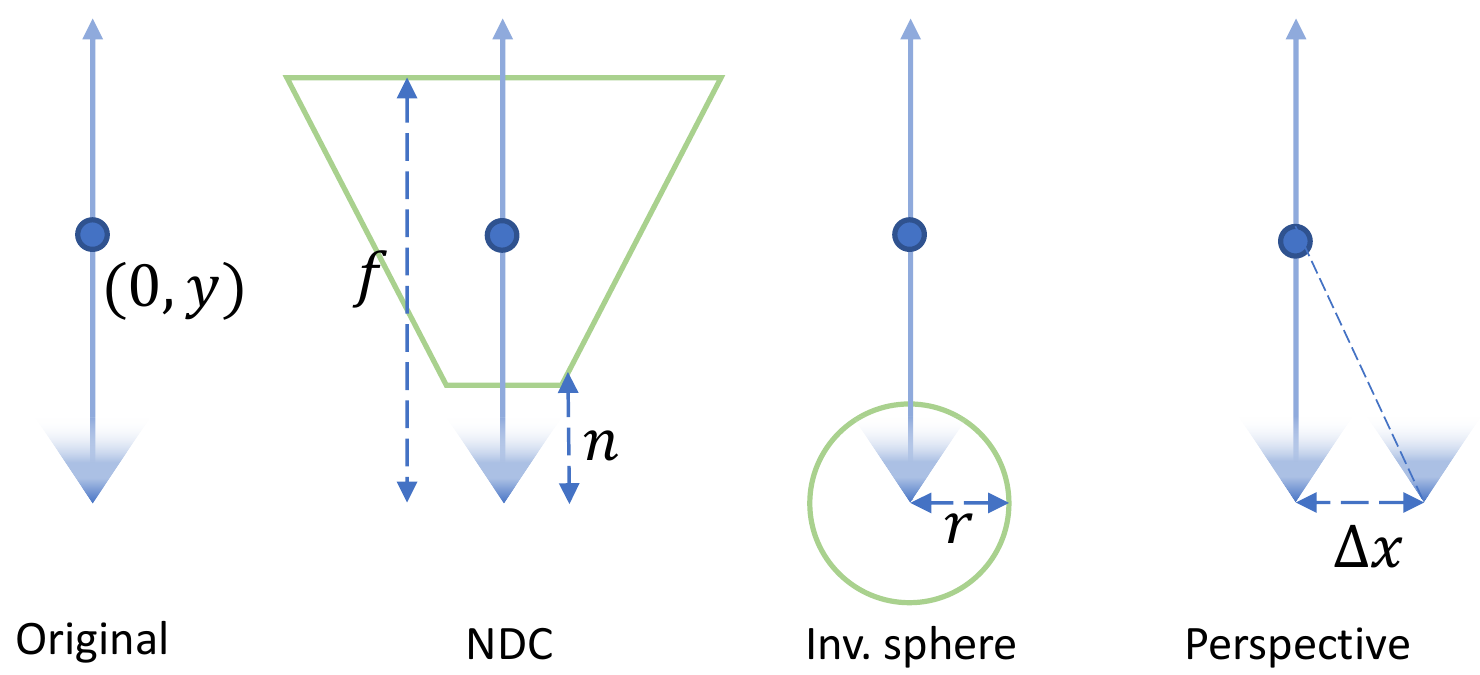}
  \caption{Different warp coordinates.}
  \label{fig:rebuttal_connection}
\end{figure}

\section{Connection to NDC warping and Inv. sphere warping} 
In the main paper, we intuitively show the connection of our proposed perspective warping to NDC warping and inverse sphere warping. Here we mathematically 
 analyze the connections using two forward-facing 1D cameras that project 2D points onto their 1D camera plane as shown in Fig.~\ref{fig:rebuttal_connection}. 
The proper perspective warping utilizes the image coordinates of two cameras as the warping coordinates. Thus, the point with coordinate $(0,y)$ in the original Euclidean space will be mapped to $(0, -\frac{\Delta_x}{y})$ in the warping space.
Meanwhile, the coordinates of this point in the NDC space and inverse sphere space are $(0, \frac{f+n}{f-n} - \frac{2fn}{f-n}\cdot\frac{1}{y})$ and $(0, 2 - \frac{r}{y})$ respectively, where $n,f$ are preset near-far depth and $r$ is preset sphere radius. When $\Delta_x=\frac{2fn}{f-n}$ or $\Delta_x=r$, the perspective warping is equivalent to NDC warping or inverse sphere warping with a constant offset. However, theoretically proving such connections in general cases of the 3D space is very complex since it involves sampling points and PCA analysis on sample points.

\section{Additional Experimental Results}
\subsection{Training for longer steps}
We provide quantitative results on training \ourmethod and instant-NGP for a longer time on the Free dataset (Table~\ref{tab:supp_longer}) and NeRF-360-V2 dataset (Table~\ref{tab:supp_longer_360}). When trained for a longer time, \ourmethod and Instant-NGP can obtain better rendering quality. As shown in Table~\ref{tab:supp_longer}, on the Free dataset, training Instant-NGP for a longer time (15m, 50k steps) does not achieve better rendering quality than \ourmethod (12m, 20k steps). Moreover, increasing the hash table size from $2^{19}$ to $2^{20}$ helps improve the performance of \ourmethod on the Free dataset (\ourmethod$_{\rm 50k-large}$).

\begin{table}[t!]
    \centering
    \begin{tabular}{@{}l@{\hskip 3pt}|@{\hskip 3pt}c@{\hskip 8pt}c@{\hskip 8pt}c@{\hskip 8pt}c@{}}
    \hline
    Method & Tr. time & PSNR{\scriptsize$\uparrow$} & SSIM{\scriptsize$\uparrow$} & LPIPS{\scriptsize(VGG)$\downarrow$} \\
    \hline\hline
    {\small Instant-NGP$_{\rm 20k}$} & 6m & 24.43 & 0.677 & 0.413 \\
    {\small Instant-NGP$_{\rm 50k}$} & 15m & 25.07 & 0.703 & 0.376 \\
    \hline
    {\small \ourmethod$_{\rm 20k}$} & 12m & 26.32 & 0.779 & 0.276 \\
    {\small \ourmethod$_{\rm 50k}$} & 30m & 26.85 & 0.811 & 0.235 \\
    {\small \ourmethod$_{\rm 50k-large}$} & 36m & 27.19 & 0.833 & 0.204 \\
    \hline
    \end{tabular}
    \vspace{-1em}
    \caption[]{{\bf Training Instant-NGP and \ourmethod for longer time on the Free dataset.}}
    \label{tab:supp_longer}
\end{table}

\begin{table}[t!]
    \centering
    \begin{tabular}{@{}l@{\hskip 3pt}|@{\hskip 3pt}c@{\hskip 8pt}c@{\hskip 8pt}c@{\hskip 8pt}c@{}}
    \hline
    Method & Tr. time & PSNR{\scriptsize$\uparrow$} & SSIM{\scriptsize$\uparrow$} & LPIPS{\scriptsize(VGG)$\downarrow$} \\
    \hline\hline
    {\small Instant-NGP$_{\rm 20k}$} & 6m & 26.24 & 0.716 & 0.404 \\
    {\small Instant-NGP$_{\rm 50k}$} & 17m & 26.55 & 0.733 & 0.382 \\
    \hline
    {\small \ourmethod$_{\rm 20k}$} & 14m & 26.39 & 0.746 & 0.361 \\
    {\small \ourmethod$_{\rm 50k}$} & 33m & 26.92 & 0.771 & 0.333 \\
    \hline
    \end{tabular}
    \vspace{-1em}
    \caption[]{{\bf Training Instant-NGP and \ourmethod for longer time on the NeRF-360-V2 dataset.}}
    \label{tab:supp_longer_360}
\end{table}

\begin{table*}[htb]
    \centering
    \begin{tabular}{l|cccccccc|c}
    \hline
    Warping method & Fern & Flower & Fortress & Horns & Leaves & Orchids & Room & Trex & Mean \\
    \hline\hline
    {NDC Warp} & {\bf 24.82} & 27.87 & {\bf 31.22} & {\bf 27.37} & 20.74 & {\bf 19.91} & {\bf 31.75} & 26.77 & {\bf 26.31} \\
    {Inv. Warp} & 24.40 & 27.40 & 30.96 & 27.19 & 20.59 & 19.72 & 31.23 & 26.66 & 26.02 \\
    {Pers. Warp} & 24.71 & {\bf 27.88} & {\bf 31.22} & 27.22 & {\bf 20.84} & 19.82 & 31.64 & {\bf 27.03} & 26.29 \\
    \hline
    \end{tabular}
    \vspace{-1em}
    \caption[]{{\bf Different warping functions on MLP-based NeRFs on the LLFF dataset.}}
    \label{tab:supp_diff_warp_llff}
    \vspace{-0em}
\end{table*}

\begin{figure}[!b]
  \includegraphics[width=\linewidth]{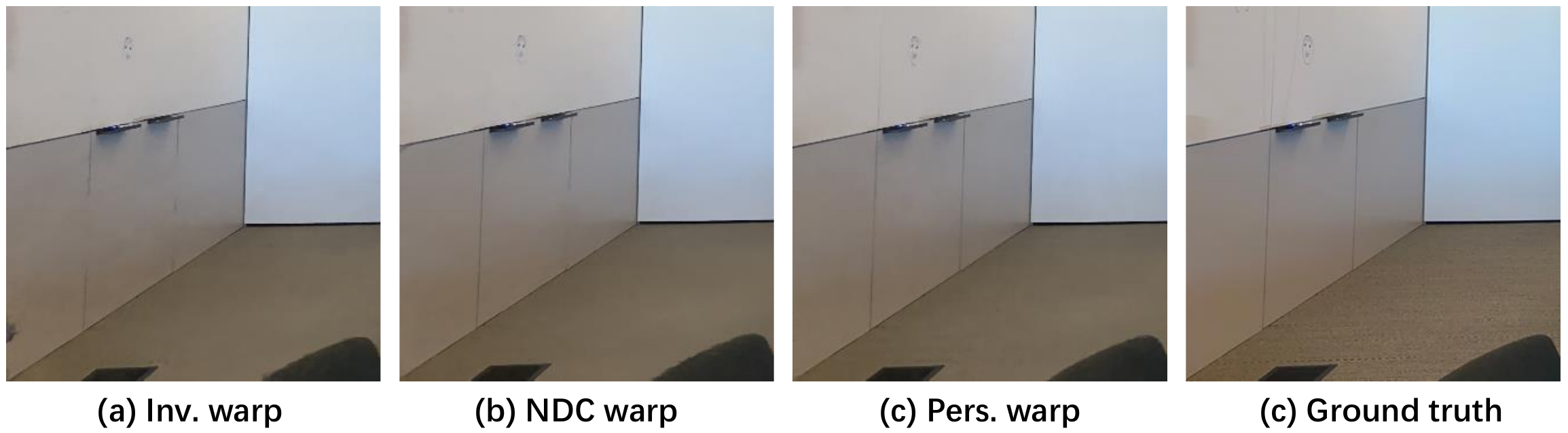} \caption{{\bf Visual comparions among different warping methods on the ``Room'' case of LLFF dataset. }}
  \label{fig:ablation_mlp_nerf}
\end{figure}

\subsection{Compatibility with MLP-based NeRF}
In this section, we provide results of applying perspective warping on MLP-based NeRF.
In this experiment, for each setting, we train a neural radiance field represented by an 8-layer fully-connected MLP for 250K steps, and use different warping functions before feeding the positional encoding to the MLP. For the perspective warping, we do not subdivide the spaces and also only use one single MLP. We also provide results of other warping functions using the same MLP-based NeRF, as shown in Table~\ref{tab:supp_diff_warp_llff}. In the forward-facing setting, our perspective warping (mean PSNR: 26.29) and NDC warping (26.31) perform better than the inverse sphere warping (26.02).
The PSNR of NDC warping is slightly worse than the reported PSNR by the original paper~\cite{MildenhallSTBRN20} due to fewer training steps (ours: 250K steps, official: 1M steps). Fig.~\ref{fig:ablation_mlp_nerf} provides qualitative results on the ``Room'' case of LLFF dataset, and our perspective warping method presents more visual details in the synthesized image, which demonstrates that the proposed perspective warping is compatible with MLP-based NeRF.

\subsection{View extrapolation}
\begin{figure}[htb]
\includegraphics[width=\linewidth]{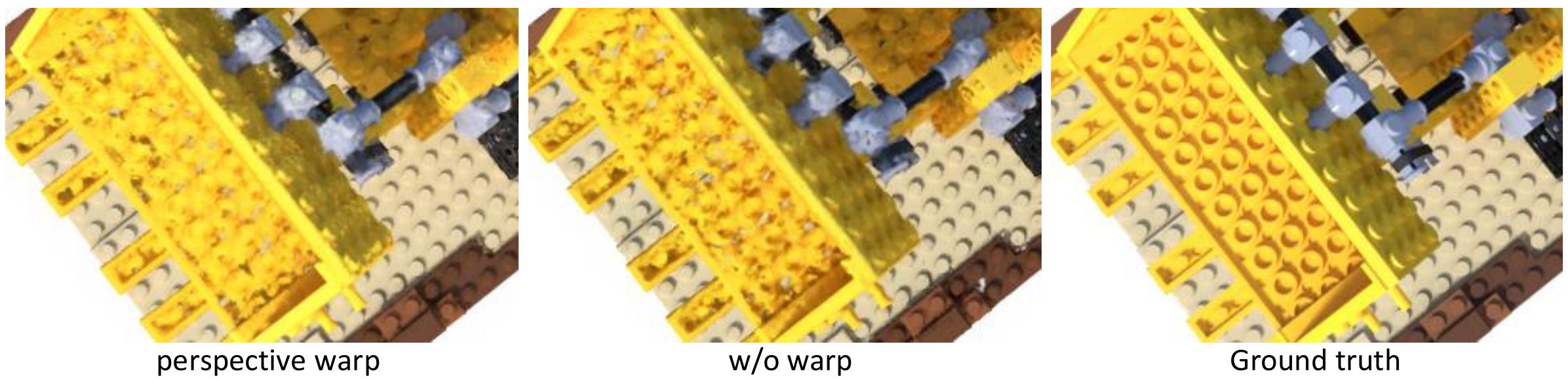}
\caption{\bf The NVS result of an extrapolated view.}
\label{fig:rebuttal_extrapolation}
\end{figure}

Here we additionally conduct an experiment for view extrapolation on the ``Lego'' case from the NeRF synthetic dataset. In this experiment, we choose images with elevation angles less than 30$^{\circ}$ for training and the others for testing. As shown in Table~\ref{tab:rebuttal_extrapolation} and Fig.~\ref{fig:rebuttal_extrapolation}, the result of our perspective warping method is similar to that using original Euclidean space.

\begin{table}[htb]
    \centering
    \begin{tabular}{@{}l@{\hskip 3pt}|@{\hskip 3pt}c@{\hskip 8pt}c@{\hskip 8pt}c@{\hskip 8pt}c@{}}
    \hline
    Elevation & $[0^\circ, 30^\circ)$ & $[30^\circ, 60^\circ)$ & $[60^\circ, 90^\circ)$ \\
    \hline\hline
    {\small Pers. warp} & 37.63 & 30.19 & 27.35 \\
    {\small w/o warp} & 37.15 & 30.35 & 27.03 \\
    \hline
    \end{tabular}
    \vspace{-1em}
    \caption[]{\bf Results on view extrapolation in the metric of PSNR.}
    \label{tab:rebuttal_extrapolation}
\end{table}

\subsection{Additional ablations}
\begin{table*}[t!]
    \centering
    \begin{tabular}{c|cccccccc}
    \hline
    Setting & Hydrant & Lab & Pillar & Road & Sky & Stair & Grass \\
    \hline\hline
    {w/o warp + Disp. sample} & 23.23 & 25.06 & 26.81 & 25.46 & 25.44 & 27.64 & 21.33 \\
    {Inv. warp + Disp. sample} & 24.05 & 25.43 & 27.09 & 26.24 & 26.27 & 27.66 & 22.34 \\
    {Inv. warp + Exp. sample} & 24.15 & 25.58 & 27.86 & 26.21 & 26.27 & 28.41 & 21.94 \\
    {Pers. warp + Exp. sample} & 24.31 & 25.79 & 28.65 & 26.60 & 26.15 & 29.08 & {\bf 22.89} \\
    {Pers. warp + Pers. sample} & {\bf 24.34} & {\bf 25.92} & {\bf 28.76} & {\bf 26.76} & {\bf 26.41} & {\bf 29.19} & 22.87 \\
    \hline
    \end{tabular}
    \vspace{-1em}
    \caption[]{{\bf Scene breakdown of our ablation studies on the warping and sampling methods.}}
    \label{tab:ablation_break_down}
    \vspace{-0em}
\end{table*}

\begin{figure}[!t]
  \includegraphics[width=\linewidth]{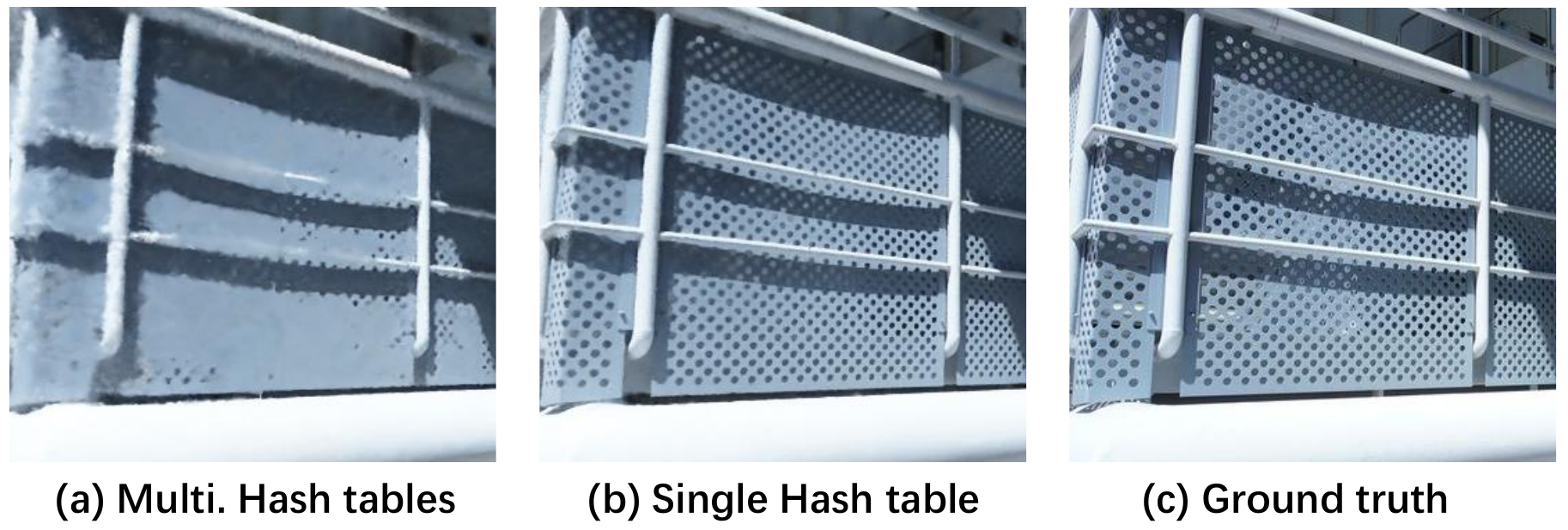} \caption{{\bf Visual comparison between using multiple hash tables (a) and single hash table (b) on the ``Sky'' case of Free dataset.}}
  \label{fig:ablation_hash_table}
\end{figure}

\textbf{Single v.s. multiple hash tables.} We test \ourmethod on the setting with multiple hash tables, i.e., one hash table for each octree node,  with the same budget of parameters as the setting of a single hash table used in the paper. In this setting, the size of each hash table is $L/n_l$, where $L=2^{19}$ is the overall table size and $n_l$ is the number of leaf octree nodes. Fig.~\ref{fig:ablation_hash_table} shows that when using multiple hash tables, the quality degrades clearly compared to the setting of using a single hash table with multiple hash functions. The reason is that using a global hash table has more flexibility in allocating the representation capacity to different regions.

\begin{figure}[!b]
  \includegraphics[width=\linewidth]{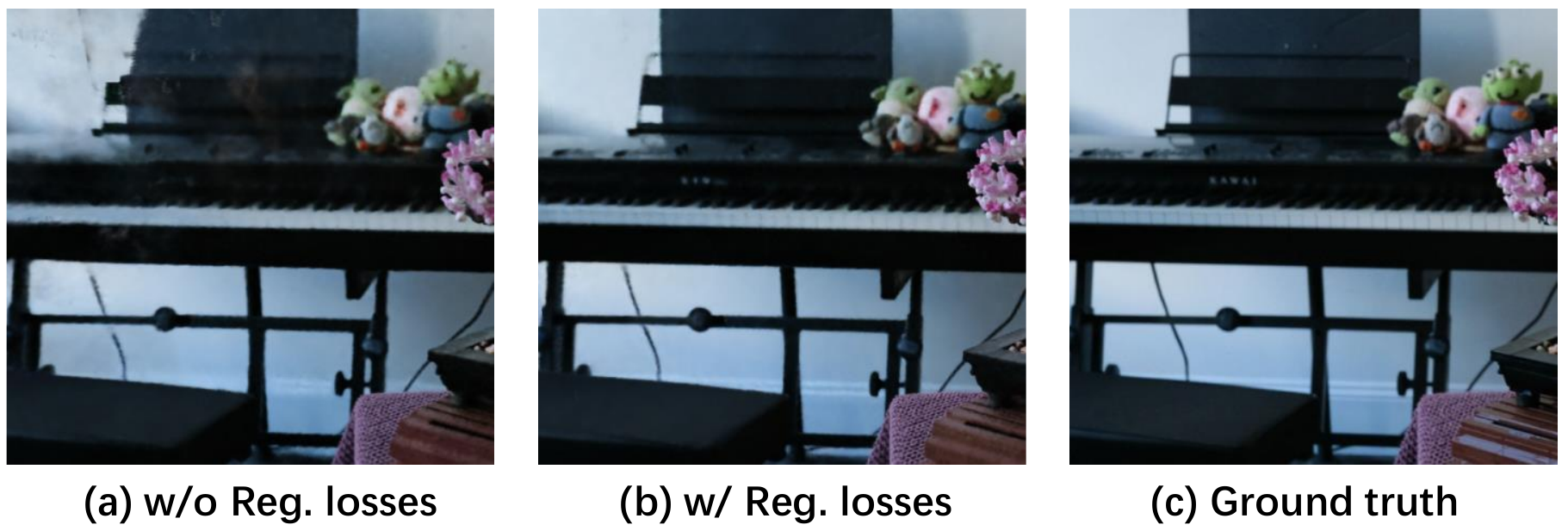} \caption{{\bf Visual comparison between w/o regularization losses (a) and w/ regularization losses (b) on the ``Bonsai'' case in NeRF-360-V2 dataset.}}
  \label{fig:ablation_reg}
\end{figure}

\textbf{Effect of regularization losses.} As shown in Fig.~\ref{fig:ablation_reg}, when regularization losses are not used, the foggy artifacts appear and the rendered result is not clear, especially on the regions with pure colors.

\begin{table*}[htb]
    \centering
    \begin{tabular}{l|ccccccc}
    \hline
    Method & Hydrant & Lab & Pillar & Road & Sky & Stair & Grass \\
    \hline\hline
    NeRF++~\cite{ZhangRSK20} & 22.21 & 21.82 & 25.73 & 23.29 & 23.91 & 26.08 & 21.26 \\
    mip-NeRF-360 & {\bf 25.03} & {\bf 26.57} & {\bf 29.22} & {\bf 27.07} & {\bf 26.99} & {\bf 29.79} & {\bf 24.39} \\
    \hline
    mip-NeRF-360 (short)~\cite{BarronMVSH22} & 21.01 & 21.17 & 24.12 & 21.49 & 22.29 & 24.27 & 19.87 \\
    Plenoxels~\cite{YuFTCR22} & 19.82 & 18.12 & 18.74 & 21.31 & 18.22 & 21.41 & 16.28 \\
    DVGO~\cite{SunSC22} & 22.10 & 23.78 & 26.22 & 23.53 & 24.26 & 26.65 & 20.75 \\
    Instant-NGP~\cite{mueller2022instant} & 22.30 & 23.21 & 25.88 & 24.24 & 25.80 & 27.79 & 21.82 \\
    \ourmethod & {\bf 24.34} & {\bf 25.92} & {\bf 28.76} & {\bf 26.76} & {\bf 26.41} & {\bf 29.19} & {\bf 22.87} \\
    \hline
    \end{tabular}
    \vspace{-1em}
    \caption[]{{\bf Scene breakdown on the Free dataset.}}
    \label{tab:compare_free_break_down}
    \vspace{-0em}
\end{table*}

\begin{table*}[htb]
    \centering
    \begin{tabular}{l|cccccccc}
    \hline
    Method & Fern & Flower & Fortress & Horns & Leaves & Orchids & Room & Trex \\
    \hline\hline
    NeRF~\cite{MildenhallSTBRN20} & {\bf 25.17} & 27.40 & 31.16 & 27.45 & 20.92 & {\bf 20.36} & {\bf 32.70} & 26.80 \\
    mip-NeRF~\cite{BarronMTHMS21} & 25.12 & {\bf 27.79} & {\bf 31.42} & {\bf 27.55} & {\bf 20.97} & 20.28 & 32.52 & {\bf 27.16} \\
    \hline
    Plenoxels~\cite{YuFTCR22} & {\bf 25.46} & 27.83 & 31.09 & 27.58 & {\bf 21.41} & 20.24 & 30.22 & 26.48 \\
    TensoRF~\cite{ChenXGYS22} & 25.27 & {\bf 28.60} & 31.36 & {\bf 28.14} & 21.30 & 19.87 & {\bf 32.35} & 26.97 \\
    DVGO~\cite{SunSC22} & 25.08 & 27.62 & 30.44 & 27.59 & 21.00 & {\bf 20.33} & 31.53 & 27.17 \\
    Instant-NGP~\cite{mueller2022instant} & 25.13 & 27.07 & 30.96 & 27.32 & 12.08 & 19.80 & 31.56 & 26.82 \\
    \ourmethod & 25.26 & 27.48 & {\bf 31.49} & 27.84 & 20.68 & 20.10 & 32.23 & {\bf 27.26} \\
    \hline
    \end{tabular}
    \caption[]{{\bf Scene breakdown on the LLFF dataset.}}
    \label{tab:compare_llff_break_down}
\end{table*}

\begin{table*}[htb]
    \centering
    \begin{tabular}{l|ccccccc}
    \hline
    Method & Bicycle & Bonsai & Counter & Garden & Kitchen & Room & Stump \\
    \hline\hline
    NeRF++~\cite{ZhangRSK20} & 22.64 & 29.15 & 26.38 & 24.32 & 27.80 & 28.87 & 24.34 \\
    mip-NeRF-360~\cite{BarronMVSH22} & {\bf 23.99} & {\bf 33.06} & {\bf 29.51} & {\bf 26.10} & {\bf 32.13} & {\bf 31.53} & {\bf 26.27} \\
    \hline
    Plenoxels~\cite{YuFTCR22} & 21.39 & 23.65 & 25.23 & 22.71 & 24.00 &	26.38 &	20.08 \\
    DVGO~\cite{SunSC22} & {\bf 22.12} & 27.80 & 25.76 & 24.34 & 26.00 & 28.33 & 23.59 \\
    Instant-NGP~\cite{mueller2022instant} & 22.08 & {\bf 29.86} & {\bf 26.37} & 24.26 & 28.27 & 28.90 & 23.93 \\
    \ourmethod & 22.11 & 29.65 & 25.36 & {\bf 24.76} & {\bf 28.97} & {\bf 29.30} & {\bf 24.60} \\
    \hline
    \end{tabular}
    \vspace{-1em}
    \caption[]{{\bf Scene breakdown on the NeRF-360-V2 dataset.}}
    \label{tab:compare_360_break_down}
    \vspace{-0em}
\end{table*}

\subsection{Per-scene results}
We provide the per-scene results on the Free dataset, NeRF-360-V2 dataset, and LLFF dataset in Table~\ref{tab:compare_free_break_down}, Table~\ref{tab:compare_360_break_down}, and Table~\ref{tab:compare_llff_break_down} respectively. The results are reported in the metric of PSNR. Table~\ref{tab:ablation_break_down} provides per-scene results on different warping and sampling methods on the Free dataset. We note that the longer the trajectory is (e.g., the ``stair'' and ``grass''), the relatively better performance our perspective warping method with the perspective sampling achieves than the inverse sphere warping method.

\begin{figure*}[htb]
  \includegraphics[width=\linewidth]{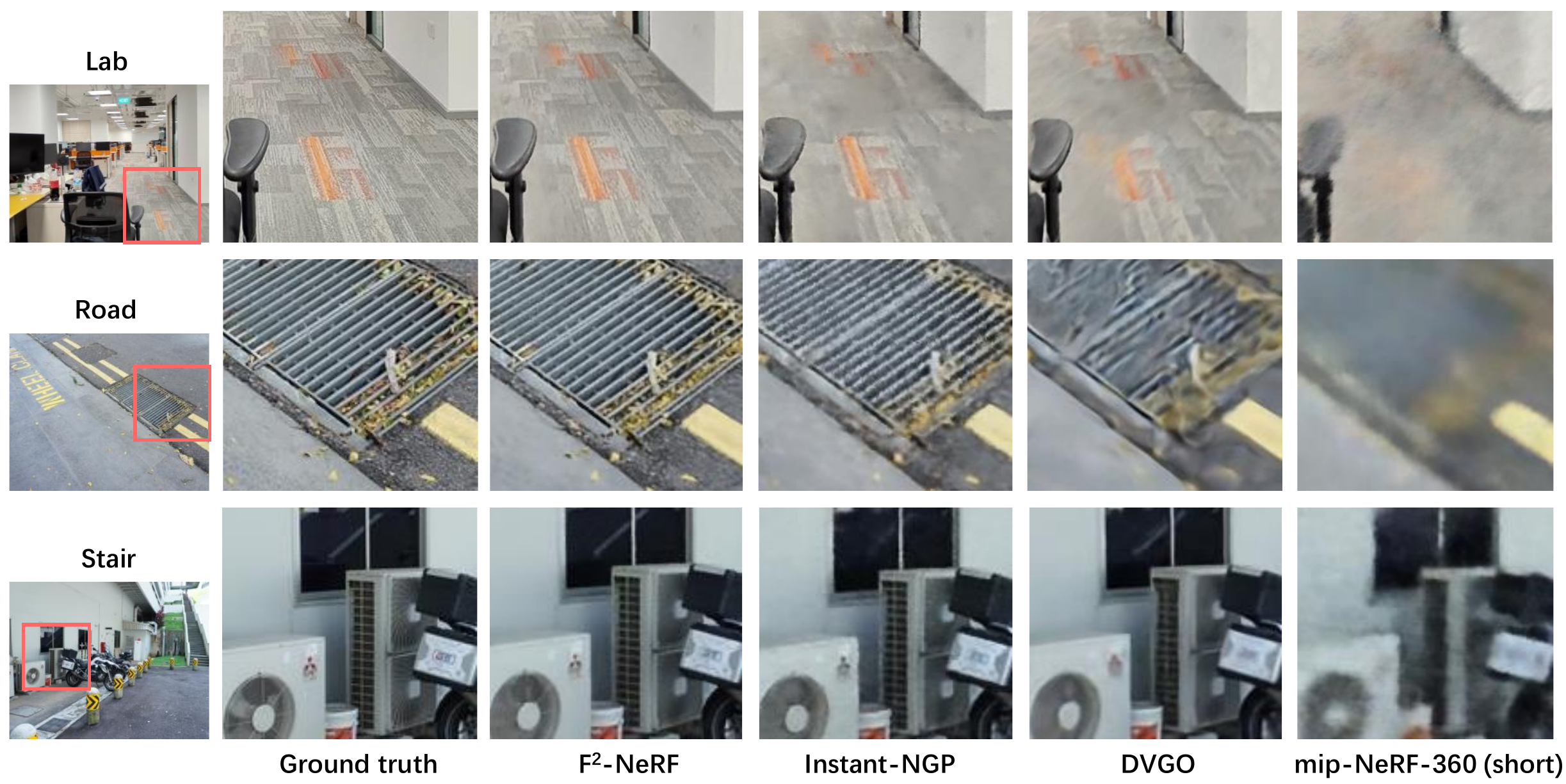} \caption{{\bf Additional visual comparions on the Free dataset.}}
  \label{fig:addtional_compare_free}
\end{figure*}

\subsection{More visual results}
We provide more visual comparisons on the Free dataset in 
Fig.~\ref{fig:addtional_compare_free}.

\end{document}